# Facial Expression Analysis under Partial Occlusion: A Survey [*1]

Ligang Zhang[1], Brijesh Verma[1], Dian Tjondronegoro[2] and Vinod Chandran[3]
[1]Central Queensland University
[2]Southern Cross University
[3]Queensland University of Technology

Automatic machine-based Facial Expression Analysis (FEA) has made substantial progress in the past few decades driven by its importance for applications in psychology, security, health, entertainment and human computer interaction. The vast majority of completed FEA studies are based on non-occluded faces collected in a controlled laboratory environment. Automatic expression recognition tolerant to partial occlusion remains less understood, particularly in real-world scenarios. In recent years, efforts investigating techniques to handle partial occlusion for FEA have seen an increase. The context is right for a comprehensive perspective of these developments and the state of the art from this perspective. This survey provides such a comprehensive review of recent advances in dataset creation, algorithm development, and investigations of the effects of occlusion critical for robust performance in FEA systems. It outlines existing challenges in overcoming partial occlusion and discusses possible opportunities in advancing the technology. To the best of our knowledge, it is the first FEA survey dedicated to occlusion and aimed at promoting better informed and benchmarked future work.

## 1 Introduction

Facial expressions of emotion are a major channel in daily human-human communication. Machine-based automatic analysis of expression from human faces is an important part of artificial intelligence capabilities. It has potential applications in various fields, such as intelligent tutoring systems, emotionally sensitive robots, driver fatigue monitoring, personalized service provision, interactive game design, and emotion based data retrieval, categorization and management. However, automatic Facial Expression Analysis (FEA) in an unconstrained real-life situation is still difficult. It encounters a variety of challenges arising factors such as occlusion, face pose variations, illumination changes, head motion, and differences in the age, gender, skin color and culture of the subject between training and testing phases of a system. An ideal FEA system should be able to handle all these challenges. While face and facial expression recognition systems have addressed most of these factors systematically, occlusion is often overlooked and assumed to be taken care of by controlled acquisition – which is only true for laboratory or prescribed acquisition conditions.

Partial occlusion presented in the face is one of the major obstacles for accurate FEA in real-world conditions. In real-life situations, there is a high likelihood that some parts of the face become obstructed by sunglasses, a hat, a scarf, hands moving over the mouth, a moustache or hair, etc. Occlusion can substantially change the visual appearance of the face and severely deteriorate the performance of FEA systems. The presence of occlusion increases the difficulty of extracting discriminative features from occluded facial parts due to inaccurate feature location, imprecise face alignment or face registration error [Ekenel and Stiefelhagen 2009]. It also introduces noise and outliers to extracted features leading to higher intra-expression variations. An FEA system with occlusion handling capacity aims to achieve accurate emotion recognition even when a portion of the face is occluded. The system can be useful in various real-life scenarios, particularly those with frequently occurring occlusion, such as students wearing glasses in online tutoring, patients wearing medical masks in medical diagnosis, and players with pose variations in game entertainment.

Although the significance of handling facial occlusion has been widely recognized in the research community for a long time, there has been a significant delay in the development of relevant algorithms and systems. In earlier surveys on FEA [Pantic and Rothkrantz 2000], [Fasel and Luettin 2003], no study reported specifically designed techniques to overcome facial occlusion. This situation remained the same in the survey on audio-visual affect recognition in 2009 [Zeng *et al.* 2009], which concluded that most human affect recognizers are evaluated using non-occluded facial data and that developing methods that are robust to occlusion is an important issue that is yet to be addressed. In recent surveys on FEA in 2012 [Bettadapura 2012], 2015 [Owusu *et al.* 2015], [Sariyanidi *et al.* 2015] and 2016 [Corneanu *et al.* 2016], several papers were cited that exploited the most informative facial parts or developed automatic systems for FEA. Current literature still lacks a comprehensive and focused survey of existing efforts in overcoming partial occlusion for FEA.

* This material is based on work supported under Australian Research Council's Linkage Project (LP140100939).
Author's address: Ligang Zhang and Brijesh Verma, Centre for Intelligent Systems, Central Queensland University, Brisbane, QLD, Australia 4000; emails: {l.zhang, b.verma}@cqu.edu.au; Dian Tjondronegoro, School of Business and Tourism, Southern Cross University, Bilinga, QLD, Australia 4225; email: dian.tjondronegoro@scu.edu.au; Vinod Chandran, Science and Engineering Faculty, Queensland University of Technology, Brisbane, QLD, Australia 4000; email: v.chandran@qut.edu.au.

This paper aims to bridge this gap. It is expected that it can serve as a good reference for developing techniques toward robust FEA in the presence of occlusion. The outline of this paper follows the concept map illustrated in Fig.1, and the remainder of the paper is organized as follows: Section 2 introduces background knowledge about FEA under partial occlusion, including its brief history, methods for representing emotions, and the major types and characteristics of facial occlusion. Section 3 examines related databases for performance evaluation of FEA systems. Section 4 reviews existing FEA approaches that can automatically recognize emotional states from occluded faces. We identify top five techniques that can be used as baselines for performance evaluations of future algorithms. Section 5 summarizes investigations on the effect of occlusion on the performance of recognizing facial expressions based on computer vision or human perception. In Section 6, we present discussions about existing challenges and possible opportunities, covering the aspects of data creation, occlusion detection, feature extraction, context information, and multiple modalities and disciplines. Finally, Section 7 draws some conclusions.

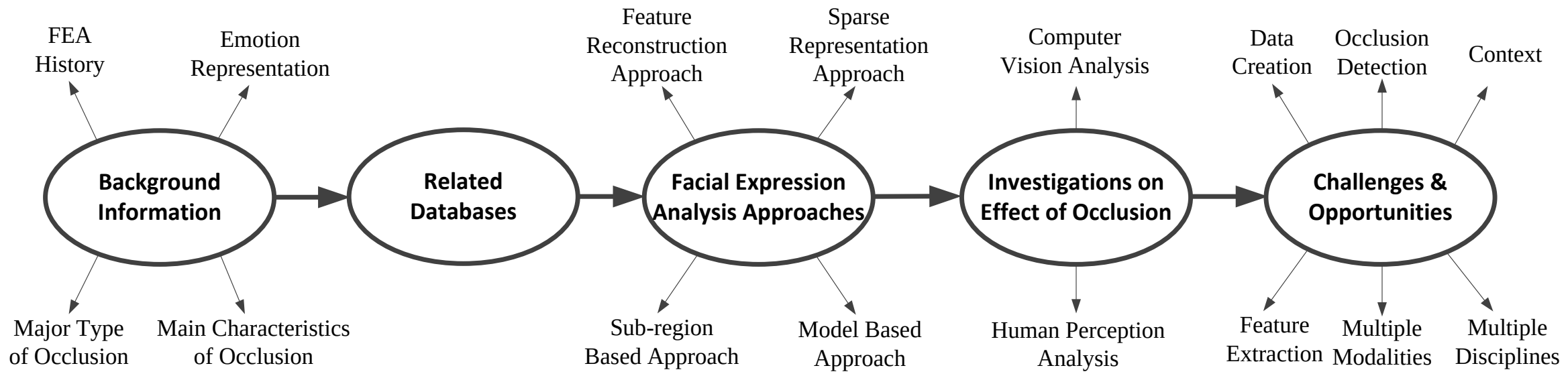

**Fig. 1. Overview of the structure of the survey on FEA under partial occlusion.**

## 2 Background

### 2.1 Brief History of Facial Expression Analysis (FEA) under Occlusion

Research on FEA can be tracked back to the study of physiognomy in the 4$^{th}$ century BC, which assessed the character or personality of a person from his/her outer appearance, primarily the face. This study was later extensively extended to explore the relationship between facial expressions and the movements of head muscles in the 17$^{th}$ century [Bettadapura 2012]. Since then, one pioneering work that has significantly impacted today's automatic FEA systems was done by Charles Darwin, who provided evidence to the existence of some basic emotions universally across cultures and ethnics. Another work was done by Ekman and his colleagues [Ekman 1978], who designed the Facial Action Coding System (FACS) to encode the states of facial expressions using facial Action Units (AUs). Until the 1980s, most FEA work was conducted by philosophers and psychologists (see a review of early work [Keltner *et al.* 2003]). Arguably, Kenade [1973] and Suwa *et al.* [1978] are two earliest investigators on recognizing facial expressions using computer technologies, and they developed computer programs to extract facial points for analyzing human faces and representing facial expressions. After relatively slow progress in the 1970s and 1980s, the 90s witnessed increased development of automatic FEA systems, moving from analyzing deliberately posed prototypical emotions of near-frontal faces collected in controlled laboratory settings to spontaneously evoked emotions collected from non-constrained settings (see reviews [Pantic and Rothkrantz 2000],[Fasel and Luettin 2003],[Zeng, et al. 2009],[Bettadapura 2012]).

On the recognition of the big impact of facial occlusion on FEA, since the 1920s, a host of psychosocial studies (e.g. Ruckmick [Ruckmick 1921], Dunlap [Dunlap 1927], Boucher and Ekman [Boucher and Ekman 1975]) have investigated the facial parts that are most important for human perception and recognition of facial expressions from partially occluded faces. However, the first machine system for FEA in the presence of occlusion was presented in 2001 by Bourel *et al.* [Bourel 2001], who recovered geometric facial points for overcoming occlusion in regions of the upper face, mouth, and left/right half of the face. Inspired by this work, most initial efforts [Bourel *et al.* 2002], [Towner and Slater 2007] focused on the recovering of geometric features from occluded faces in static images and the classification of facial expressions using a single global classifier. More recent advancements [Xia *et al.* 2009], [Cotter 2011], [Liu *et al.* 2014c] have shifted to the adoption of texture features and their combination with geometric features in temporal 2D or 3D video sequences, as well as the fusion of multiple local classifiers from different facial regions to derive a final classification decision for the whole face. Kotsia *et al.* [2008] presented, to our best knowledge, the most comprehensive analysis on the impact of facial occlusion on the recognition of six basic emotions based on FER systems. The results were found to agree in overall with those from human observers. With the popularity of deep learning techniques, recent studies [Cheng *et al.* 2014], [Tősér *et al.* 2016], [Batista *et al.* 2017] have focused on the use of deep neural networks to directly perform FEA on occluded facial images without involving steps of occlusion detection, hand-engineering feature extraction, and classifier design.

Although more types of occlusion and datasets have been included in recent studies, most of existing studies are primarily based on a limited number of artificially generated types of occlusion and the present progress is relatively slow.

## 2.2 Methods for Representing Facial Expressions

Representing facial expressions is a prerequisite for evaluating the effectiveness of FEA systems, particularly in the presence of occlusion. Facial expressions are generally represented using two methods in exiting studies: message based and movement of facial components based.

The message based method can be further divided into discrete categorical and continuous dimensional methods. The discrete categorical method is perhaps the most long-standing and widely used way for describing facial expressions by psychologists. In this method, an expression is assigned to one of pre-defined prototypical categories, including six basic emotional states - *anger (AN), disgust (DI), fear (FE), happiness (HA), sadness (SA),* and *surprise (SU)* [Ekman 1994], and non-basic emotional states such as depression, agreement, distress and disappointment . This intuitive theory creates a convenient way of representing observed facial expressions in daily lives using a list of emotion words with concrete meaning, which is highly consistent with human understanding. However, the theory can represent only a small portion of possible complicated and mixed emotions in natural communication conditions. Studies [Maja *et al.* 2005] have shown that "pure expressions of prototypical emotions are less frequently elicited and blends of emotional displays are often shown by humans in real-life situations".

The continuous dimensional method was derived from the field of psychology [Russell 1980]. It describes facial expressions using continuous axes in a multiple dimensional space and represents each expression as a point or a region in the space. The most commonly used spaces are composed of two- and three- dimensional representations, such as activation or arousal, valence, power, and expectancy. The advantage of dimensional spaces over discrete categories and AUs lies in the use of a set of continuous axis values to represent a large amount of different types of emotions, including those naturalistic non-prototypical ones that often occur in realistic conditions. They can provide useful insights into the intensity of emotions, as well as the similarity and contrast between categorical emotions. However, as the dimensional space is not intuitive, it requires specially trained annotators in emotion labeling using continuous axes. In addition, some emotions become indistinguishable in a limited number of dimensions and it is not straightforward in directly applying the analysis results into practical applications.

The movement of facial components based method uses the movements of individual facial muscles to encode facial expression states. Examples of this method include FACS [Ekman 1978], Emotional Facial Action Coding System (EMFACS), MAXimally discriminative facial movement coding system (MAX) [Izard et al. 1979], and probability-based AU space [Zhao *et al.* 2016]. The FACS, which was originally developed by Paul Ekman and his colleagues in 1978, defines a total of 44 AUs to encode movements of facial muscles. Each AU corresponds to a contraction or relaxation of individual or multiple facial muscles that can generate a certain facial action with regard to its location and intensity. A revised version was published in 2002 [Ekman et al. 2002] comprising of 32 AUs. One advantage of AUs is that a combination of relatively few AUs can effectively represent a large number (e.g., thousands) of expression states and subtle facial signals such as wink and frown. Thus, AUs are able to represent a wide variety of emotions and are suitable for describing, modeling and analyzing real-life facial expressions. In addition, AUs are objective descriptors and thus they are independent of human interpretation, which consequently reduces the subjectivity of emotion labeling. However, the challenge lies in the difficulty to accurately choose a set of relevant AUs and their combination for arbitrary naturalistic emotions, and vice versa it is also challenging to translate emotion related AUs into affective meanings. The EMFACS, as a selective application of FACS scoring, focuses on scoring only facial actions that are likely to have emotional significance. The MAX was designed to code discrete emotional states such as interest, joy, surprise, contempt, and physical distress or pain, based on a set of facial movement formulas. Rather than using binarized AU occurrence in FACS, EMFACS and MAX, the probability-based AU space treats each basic AU as an individual dimension and uses continuous coordinates on an AU axis to represent the probability of this AU occurring on a face. A hyperplane can be constructed to divide the space into regions of emotions or affective states. The space has the advantages of not requiring manual labelling of AUs from skilled experts and being more robust for AU detection in margin areas.

Another category of facial component movement based methods focuses on the recognition of Micro-Expressions (MEs). MEs are brief, involuntary facial expressions which reveal hidden emotions and are important for understanding humans' deceitful behaviours. Unlike general facial expressions, MEs are very short (i.e., 1/25 to 1/3 second), involve subtle muscle movements and are difficult to control through one's willpower. It is still a difficult task, even for humans, to precisely recognize MEs in real-life environments. Although MEs have been given relatively less attention compared to general expressions, various types of tools and systems have been developed for recognizing MEs, such as the Micro Expression Training Tool developed by Ekman [Ekman 2003], the MR recognition system [Wu *et al.* 2011], and the ME analysis system (MESR) [Li *et al.* 2017b]. Encouragingly, some systems such as MESR, were reported to outperform humans in ME recognition. One possible benefit of using MEs for FEA under occlusion is that, even a part of the face is occluded intentionally by a subject to hide his/her emotions, the true emotional state may still can be automatically revealed by recognizing those MEs hidden in unoccluded parts of the face.

## 2.3 Major Types and Characteristics of Facial Occlusion

Due to the complexity and variability of specific environments where the face presents, the types of facial occlusion that occur may vary significantly. Generally, there are two major types of partial occlusion: systematic and temporary [Towner and Slater 2007]. Systematic occlusions are caused by the existence of individual facial components (e.g., hair, mustache, or a scar), or by people wearing adornments (e.g. glasses, clothes, a hat or surgical mask, or mark-ups) as displayed in Fig. 2 (a). Temporary occlusions arise from a portion of the face being temporarily obscured by other objects (e.g., people moving across the face or hands covering the face), or from environmental condition changes (e.g., lighting and shadows), or self-occlusion due to changes in head pose (e.g., out-of-plane pose variations) or temporarily placing objects in front of the face as shown in Fig. 2 (b). Due to the necessity of constantly interacting with the environment, self-occlusion might occur more frequently than other types of temporary occlusions in daily lives. Facial occlusion is not necessarily restricted to be either systematic or temporary; instead, it can be composed of multiple types of occlusion as shown in Fig.2(c). In special cases as shown in Fig. 2 (d), blurring, pixellation, artificial masks or texts are added specifically into the face to hide personal identity or provide helpful information.

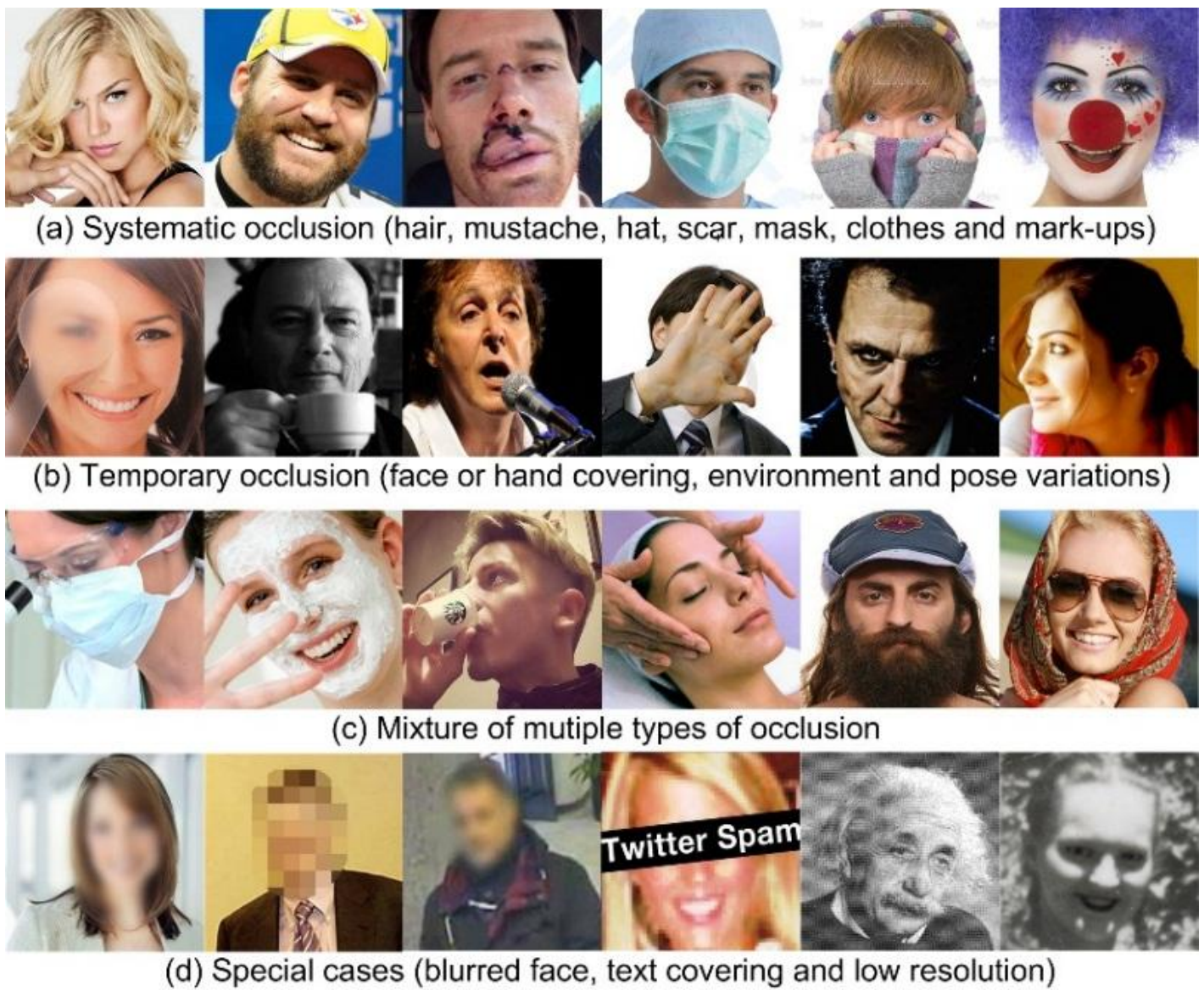

**Fig. 2. Real examples with different types of facial occlusion.**

Different from other challenges, such as pose and illumination variations whose main characteristics and the associated impact can be inferred beforehand, facial occlusion has several distinguishing characteristics that make it particularly difficult to be handled:

1) *Varying type.* The type of occlusion may vary significantly dependent on the situation where the face presents. Unless there is prior knowledge about what type of occlusion is going to occur in a specific context, FEA systems should consider all possible types, which is very challenging and technically infeasible at least for now.

2) *Mixed type.* Multiple types of occlusion may co-exist in the face. The existence and prevalence of mixed occlusion increases the difficulty of handling occlusion and investigating the effect of individual occlusion because the separation of mixed occlusion is still an unexplored field.

3) *Non-fixed location.* Most types of occlusion are generally not fixed to a certain location of the face. Although the location is roughly predictable for some systematic occlusion, such as glasses and a hat, it is still difficult to predict their precise locations and it is more difficult for temporary occlusion such as hand covering.

4) *Varied duration.* Different types of occlusion may exist for a different length of duration. Occlusion due to a hand moving over the face is anticipated to last for only few seconds, while occlusion of sunglasses often exist for the total duration of the video data. Thus, the duration of occlusion is largely dependent on the nature type of the occlusion.

5) *Unpredictable property.* Due to the variability of objects leading to occlusion, the visual properties (e.g. shape, appearance and size) of the resulting occlusion are often unpredictable. For example, there might be big differences in the visual properties between sunglasses, vision correction glasses, and protective glasses.

6) *Local impact.* Unlike pose and illumination variations, which often impact the holistic facial region, most types of occlusion impact only a small portion of the face. This property can be somehow treated as a merit because the effect of occlusion can be compensated using un-occluded facial parts.

It should be noted that in specific cases, it might be possible to roughly predict the parameters (e.g. type, location, shape, appearance and time) of occlusion that is likely to occur. For instance, a person's face is highly likely to be partially and temporally occluded by a cup and moving hands when he/she is drinking coffee. Prior knowledge about the occlusion is critical in developing techniques to overcome its impact.

## 3 Related Databases

Being able to access public facial expression datasets with well-annotated occlusion is a pre-requisite in evaluating FEA systems. Many databases [Patil *et al.* 2015] with facial occlusion were generated for face recognition, but very few have been created specifically for FEA. This section introduces related public databases (samples shown in Fig. 3) that either are widely used in existing studies with artificially superimposed occlusion, or can be potentially used for future system evaluations with naturally occurring occlusion. It is worth mentioning that there are also recently released datasets such as AFEW [Dhall *et al.* 2012], QUTFER [Zhang *et al.* 2014a] and BAUM-1 [Zhalehpour *et al.* 2016] that contain a certain level of realistic facial occlusion caused by factors such as pose variations and lighting changes, but as these datasets are not specifically created for evaluating FEA under occlusion and only limited types of occlusion occur occasionally in few cases, they are not covered here. Readers are referred to [Zafeiriou *et al.* 2016] for a survey of FEA databases collected in a wild environment.

1) The JApanese Female Facial Expression (JAFFE) database [Lyons *et al.* 1998] is a widely used benchmark dataset for FEA in early work and it includes 213 images of six basic emotions plus neutral that were posed by 10 Japanese female subjects. Each subject has three or four non-occluded frontal face images per expression and the face is roughly located at the center of the images. The images have a resolution of 256×256 pixels and have been rated on seven emotion categories by 60 subjects. Although the JAFFE database has been used less frequently in recent studies, it is still a popular dataset for evaluating the effect of artificially superimposed occlusion on the FEA.

2) The Cohn-Kanade (CK) database [Kanade *et al.* 2000] is a popular comprehensive dataset for facial expression benchmark tests. It is composed of 486 video sequences collected from 97 subjects with neutral to target displays in each sequence. The frames have a resolution of 640×480 or 640×490 pixels and are fully FACS coded. Annotation of six basic expressions has also been provided. The extended version (CK+) [Lucey *et al.* 2010] includes 593 posed expression sequences from 123 subjects and 122 spontaneous smile sequences from 66 subjects. Similar to JAFFE, CK and CK+ do not include occluded faces.

3) The BeiHang University Facial Expression (BHUFE) database [Yu-Li *et al.* 2006] includes 1,600 color videos for the frontal and 30-degree profile of 25 facial expressions from 32 college students aged 21 to 25. Each video lasts about 6 seconds and has a frame resolution of 480×640 pixels. The database includes four typical complex facial expressions: smile while hands obscure the face, anger while hand obscure the face, smile while talking, and anger while talking.

4) The Caltech Occluded Faces in the Wild (COFW) dataset [Burgos-Artizzu *et al.* 2013] comprises of 1,007 facial images collected from a variety of real-world sources by four people. It was designed to evaluate the performance of face landmark algorithms in realistic conditions, which have heavy occlusion and large shape variations that arise from the differences in expression or head pose, using accessories (e.g. hats and sunglasses), and interacting with objects. There are substantial variations in the type of occlusion and different degrees of occlusion in the faces. The average face occlusion is over 23%. The occluded/unoccluded state in all images was hand annotated.

5) The Acted Facial Expressions in the Wild (AFEW) database [Dhall, et al. 2012] is a dynamic temporal facial expressions data corpus that comprises of a total of 957 audio-visual clips extracted from 37 movies. The clips are searched by their subtitles using a list of expression keywords such as 'laughs', 'smiles', 'scared', etc. The clips are labeled with six basic emotions and actor information by human observers. The Static Facial Expressions in the Wild (SFEW) database [Dhall *et al.*] is a subset of the AFEW database, and it is composed of static frames selected from AFEW video clips. The SFEW includes 700 images labelled with six basic emotions. The images have unconstrained facial expressions with variations in pose, subject age and image resolution, as well as real-life occlusions such as glasses, eye mask, beard and hat. The AFEW (or SFEW) has been used as a benchmark dataset in the Emotion recognition in the Wild (EmotiW) challenge from 2013 to 2017 [Dhall *et al.* 2017].

6) The HAPpy PeoplE Images (HAPPEI) database [Dhall *et al.* 2013] is created to evaluate the happiness intensity of a group of people. It includes 4,886 images that are collected from Flickr by searching keywords associated with groups of people and events, such as 'party + people', 'marriage', 'reunion', 'bar', etc. All images have more than one subject and are annotated with group-level mood intensity (from neutral to thrilled). In addition, 8,500 faces in these images are also annotated for six intensities of happiness (i.e., neutral, small smile, large smile, small laugh, large laugh and thrilled), and three intensities of occlusion (i.e., face visible, partial occlusion and high occlusion). The dataset has been used as the benchmark dataset for group-level emotion recognition in the EmotiW 2016 [Dhall *et al.* 2016b]. Instead of focusing on only happiness in the HAPPEI database, the Group Affect database [Dhall *et al.* 2015b] covers images of a group of people with three emotion categories of positive, negative, and neutral. It has been used in the EmotiW 2017 [Dhall, et al. 2017].

Although partial occlusion such as beard, glasses and hat frequently present in some images, no annotation labels of the occlusion are provided in the database.

7) The 'in-the-wild' database [Zafeiriou, et al. 2016] is a recently developed dataset for facial expression analysis in naturalistic conditions. It comprises of more than 500 video collected from Youtube with people reacting to different emotional scenarios including performing an activity, a practical joke, a positive surprise, a particular video, etc. The facial responses have been annotated with arousal and valance values by three rates using the FeelTrack tool. In addition, the database also includes more than 10,000 facial images from more than 2,000 people that were collected by performing a tag based search in Google Image using keywords such as anger, feeling, fear, and pain. The facial images were annotated with 16 AUs by a trained AU coder. This database contains various types of natural occlusion caused by hands, pose variations, lighting changes, etc., but no ground truth data regarding the presence/state of occlusion was provided.

8) The Bosphorus database [Savran *et al.* 2008] is a large multi-expression and multi-pose 3D face dataset with different types of real-life face occlusions. It comprises of 4,652 face scans from 105 subjects mostly aged between 25 and 35. Facial expressions were encoded using 28 AUs and six basic emotions, and occlusion of the eyes and mouth were formed naturally by subjects rubbing their eyes, or putting hands over their mouths. Occlusion of eyeglasses was generated by asking each subject to wear a pair of eyeglasses chosen from a pool of different eyeglasses. The dataset also includes a small portion of facial images with partial occlusion by long hair.

9) The University of Cambridge 3D (Cam3D) multimodal corpus [Mahmoud *et al.* 2011] was specifically collected to analyze hand-over-face gestures. It includes 108 audio/visual segments of spontaneous facial expressions and hand-over-face gestures from 16 participants with 12 natural mental states, such as thinking, unsure, happy, surprise, anger, frustrated and confused. The data was captured using three sensors of HD cameras, Microsoft Kinect and microphones, and has a frame resolution of 640×480 or 720×576 pixels. The emotional state was annotated using crowd-sourcing techniques, and the hand-over-face gestures were annotated into three states of *action*, *hand shape* and *facial region occluded.*

10) The University of Milano Bicocca (UMB) 3D database [Colombo *et al.* 2011] consists of 1,473 2D color images and 3D depth images collected from 143 subjects. Each subject has at least four facial expressions (neutral, smiling, anger and bored) and occluded faces by scarf, hat or hands in random positions. Most subjects also have partial occlusion arising from eyeglasses, holding phones, hair, or other miscellaneous objects. 42% of the face area is occluded on average, with the largest coverage of about 84%. Each acquisition was described by labels such as occluded or non-occluded, occluding object (if any) and facial expression. In total, there are 578 occluded faces with an image resolution of 640×480 pixels. One drawback of this database is that all subjects were asked to keep their eyes closed during recordings.

11) The Binghamton University 3D Facial Expression (BU-3DFE) [Lijun *et al.* 2006] database includes both prototypical 3D facial expression shapes and 2D facial textures of 2,500 models of 100 subjects. For each shape model, the texture images were captured at two views of approximately +45 and -45 degrees. There are six basic emotions plus neural with four levels of emotion intensity. The BU-3DFE was later extended to a high-resolution 3D dynamic database – BU-4DFE [Yin *et al.* 2008]. The BU-4DFE includes 606 3D facial expression sequences from 101 subjects with 60,600 frame models in total. Each subject has six model sequences and each sequence shows one of six basic emotions. The databases are useful for simulating self-occlusion by rotating the 3D model by certain degrees of yaw, pitch, or roll.

12) The Binghamton Pittsburgh 4D spontaneous expression Database (BP4D) [Zhang *et al.* 2014c] was collected to analyze facial actions that are not deliberately posed. It contains a total of 328 sequences of high-resolution 3D images plus 2D texture videos of 1040×1392 pixels. There are eight categorical emotions, including six basic emotions, embarrassment and pain, from 41 participants (23 women and 18 men). These emotions were elicited from eight tasks. For each sequence, three types of meta data are provided, including 27 manually annotated AUs, automatically tracked head pose (pitch, yaw, and roll), and 83 2D/3D facial landmarks. It is noted that the dataset in the Facial Expression Recognition and Analysis (FERA) 2017 challenge [Valstar *et al.* 2017] was derived from the 3D model of the BP4D database. The dataset comprises of 2,952 videos for training, 1,431 videos for validation and 1,080 videos for test. The challenge focuses on the recognition of 10 frequently occurring AUs and the estimation of six intensity levels (i.e., 0-5) of seven AUs. Nine different face orientations were also created by rotating 3D sequences by -40, -20 and 0 degrees pitch and -40, 0 and 40 degrees yaw from a frontal pose.

Table 1 lists main characteristics of related databases. We can see that all databases:

- primarily use discrete categories or AUs to represent facial expressions (except 'in-the-wild');
- are largely collected in a laboratory environment (except COFE and 'in-the-wild');
- contain both artificially posed and spontaneously elicited;
- focus on occlusion by hands, glasses and hair;
- focus on self-occlusion by head pose variations (3D databases).

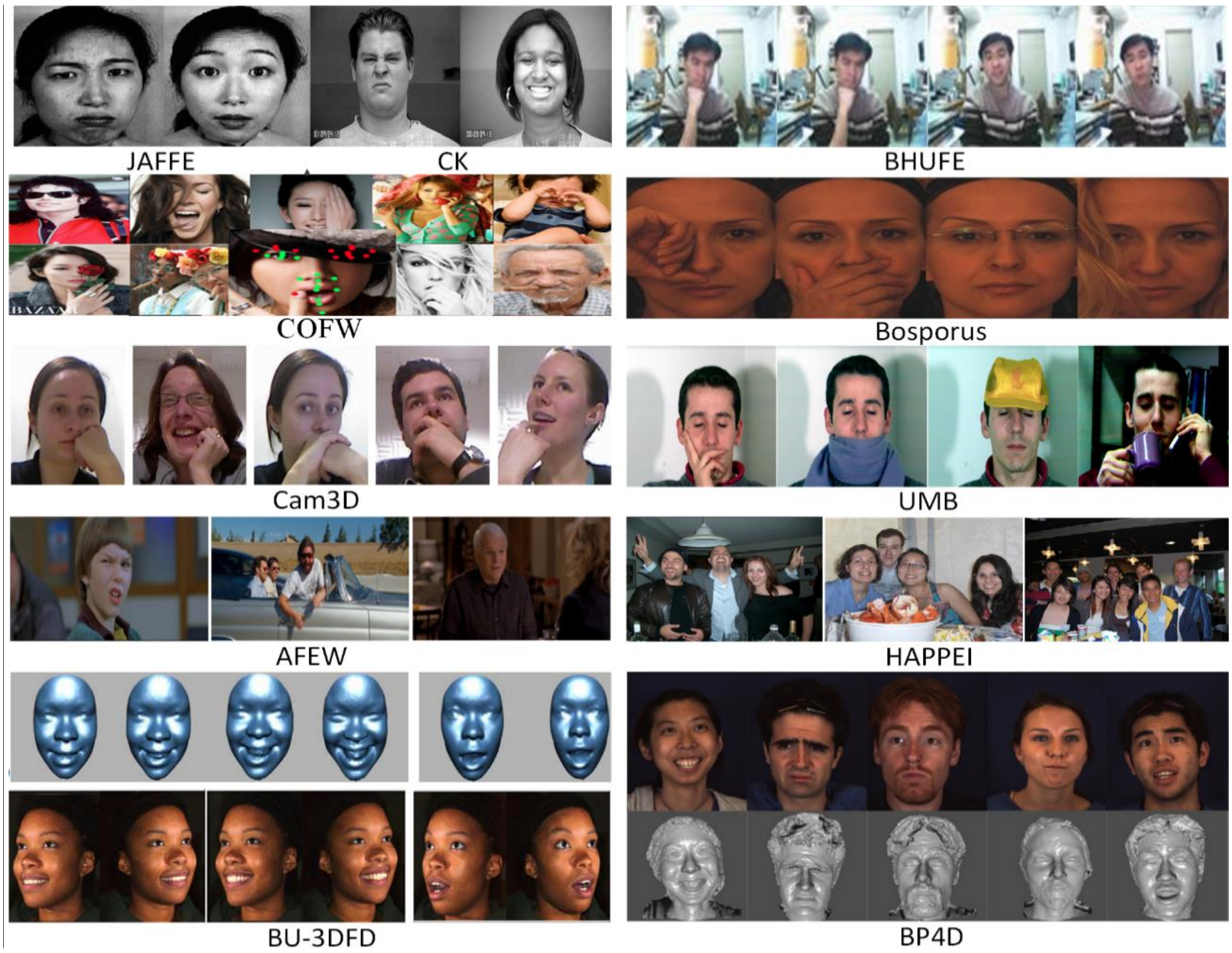

**Fig. 3. Facial samples from existing related databases.**

Table 1. Existing Databases for Validating FEA Algorithms under Facial Occlusion.

| Database | Occlusion | | Facial Expression | | Data | | | |
|---|---|---|---|---|---|---|---|---|
| | Type | A/N | Category | P/S | Size | Subject | 2D/3D | Env |
| JAFFE[1] [Lyons, et al. 1998] | - | A | six basic, neutral | P | 213 I | 10 | 2D | lab |
| CK[2] [Kanade, et al. 2000] | - | A | six basic, AUs | S | 486 V | 97 | 2D | lab |
| CK+[2] [Lucey, et al. 2010] | - | A | six basic, AUs | PS | 593 V, 122 V | 123 | 2D | lab |
| BHUFE[3] [Yu-Li, et al. 2006] | hands | N | smile, anger | P | 1,600 V | 32 | 2D | lab |
| COFW[4] [Burgos-Artizzu, et al. 2013] | sunglass, hat, food, hands, hair etc. | N | various emotions (e.g. HA) | S | 1,007 I | - | 2D | real |
| In-the-wild [Zafeiriou, et al. 2016] | hands, hat, pose, lighting etc. | N | arousal, valance, 16 AUs | S | +500 V, +10,000 I | - +2,000 | 2D | real |
| AFEW/SFEW[5] [Dhall, et al. 2012] | glasses, eye mask, beard, hand, hat etc. | N | six basic | S | 957 V, 700 I | 330 | 2D | real |
| HAPPEI [Dhall, et al. 2013] | glasses, beard, people standing front, hat etc. | N | HA (six intensities) | S | 4,886 I | - | 2D | real |
| Bosphorus[6] [Savran, et al. 2008] | hands, eyeglasses, hair, beard, moustache | N | six basic, 28 AUs | P | 4,652 scans | 105 | 3D | lab |
| Cam3D[7] [Mahmoud, et al. 2011] | hands | N | 12 mental states, e.g. thinking, unsure, happy. | S | 108 V | 16 | 3D | lab |
| UMB[8] [Colombo, et al. 2011] | Scarf, hat, hands, eyeglasses, phones, hair | N | neutral, smile, bored, hungry | P | 1,473 I | 143 | 2D/3D | lab |
| BU-3DFE[9] [Lijun, et al. 2006] | head pose variations | N | six basic, neutral | P | 2,500 models | 100 | 2D/3D | lab |
| BU-4DFE[9] [Yin, et al. 2008] | head pose variations | N | six basic, neutral | P | 60,600 models | 101 | 2D/3D | lab |
| BP4D[9] [Zhang, et al. 2014c] | head pose variations | N | eight emotions, 27 AUs | S | 328 V | 41 | 2D/3D | lab |

Note: '-' means not available.

Abbreviations: A- Artificially imposed; N – Naturally occurring; P – Posed emotion; S – Spontaneous; I – Image; V – Video.

1: http://www.kasrl.org/jaffe.html; 2: http://www.pitt.edu/~emotion/ck-spread.htm;

3: http://www.ee.buaa.edu.cn/oldeeweb/html/zykj/teachers/mx/news/7.html; 4: http://www.vision.caltech.edu/xpburgos/ICCV13;

5: https://cs.anu.edu.au/few/; 6: http://bosphorus.ee.boun.edu.tr/default.aspx; 7: http://www.cl.cam.ac.uk/research/rainbow/projects/cam3d;

8: http://www.ivl.disco.unimib.it/umbdb; 9: http://www.cs.binghamton.edu/~lijun/Research/3DFE/3DFE_Analysis.html.

## 4 Automatic Facial Expression Analysis Approaches

Rather than providing a comprehensive survey on all previous approaches on automatic FEA, this paper limits its attention to only those that have used or investigated face data with partial occlusion. For exhaustive surveys on the past approaches on FEA and affect recognition in non-occluded faces, as well as face recognition or human detection under occlusion, readers are referred to the following work:

- Introduction of early FEA approaches [Pantic and Rothkrantz 2000], [Fasel and Luettin 2003].
- Surveys of recent 2D/3D FEA methods [Bettadapura 2012], [Sandbach *et al.* 2012], [Owusu, *et al.* 2015], [Sariyanidi, *et al.* 2015], [Corneanu, *et al.* 2016].
- Summaries of affect recognition work using audio, visual, text or physiological modalities [Zeng, *et al.* 2009], [Calvo and D'Mello 2010].
- Reviews of dimensional affect analysis methods using audio, visual, and biological modalities [Gunes *et al.* 2011], [Gunes and Schuller 2013].
- Overviews of face recognition or human detection methods under partial occlusion [Azeem *et al.* 2014], [Nguyen *et al.* 2016].
- A survey of face detection methods in a wild environment [Zafeiriou *et al.* 2015].
- A survey of FEA methods using FACS [Martinez *et al.* 2017].
- A brief review of FEA using deep learning methods [Zafeiriou, *et al.* 2016].

According to the strategies used for handing facial occlusion, existing FEA approaches can be roughly divided into feature reconstruction approach, sparse representation approach, sub-region based approach, statistical model based approach, 3D data based approach, and deep learning approach. Table 2 provides an overview of these approaches with respect to occlusion detection, feature, classifier, emotion category, occlusion type, occlusion simulation, and performance. Table 3 summarizes main characteristics of these approaches regarding their pros and cons. It should be noted that sparse representation and sub-region approaches are often used for static image based FEA, while model based approach is often employed for temporal video based FEA. The feature reconstruction, 3D data based, and deep learning approaches can be used for both types of analysis.

Table 2. Summary of Six Categories of FEA approaches in the Presence of Occlusion.

| | Ref. | Algorithm | | | Database | | Occlusion | | Acc. (%) |
|---|---|---|---|---|---|---|---|---|---|
| | | Ocu. Det. | Feature | Classifier | Emotion | Size | Simulation | Type | |
| Feature reconstruction | [Bourel 2001] | N | Spatio-temp point vector | RW-KNN | AN, HA, SA, SU | CK: 100 V | missing region | upper/left/right face, mouth | >80(except SA) |
| | [Bourel, et al. 2002] | N | discrete state of point | RW-KNN | 6 basic | CK: 300 V | missing region | upper/left/right face, mouth, noise | >80(except mouth) |
| | [Towner and Slater 2007] | N | point coordinate | SVM | 6 basic | CK: 376 V | missing point | upper/lower face | 70, 82 |
| | [Kapoor *et al.* 2003] | N | point coordinate | SVM | 5 AUs, NE | natural: 80 I | real-life | N/A | 69 (each AU)<br>63 (all AU) |
| | [Zhang *et al.* 2015] | N | point coordinate | SVR+ANN | 18 AUs, 6 basic, NE, contempt | CK+: 250 I | black mask, real-life | eyes, mouth, upper/lower face | N/A |
| | [Xia, et al. 2009] | SD+ RPCA | Haar-like | Adaboost | 6 basic | JAFFE & BHUFE: 1200 I | real-life | hand, hair, sunglass | N/A |
| | [Jiang and Jia] | N | Eigen-/Fisher face | NN/SVM | 6 basic, NE | JAFFE: 213 I | black bar | mouth, eyes | 73, 78 |
| | [Cornejo *et al.* 2015] | N | Gabor/point coordinate | SVM/KNN | 6 basic, NE | JAFFE: 213 I<br>CK+: 593 I<br>MUG: I | black mask | left/right eye, eyes, bottom left/right face, bottom face | 98 (JAFFE)<br>99 (CK+)<br>99 (MUG) |
| | [Cornejo and Pedrini 2016] | N | CENTRIST | SVM | 6 basic, NE | JAFFE: 213 I<br>CK+: 593 I | black mask | | 92 (JAFFE)<br>90 (CK+) |
| Sparse representation | [Cotter 2010a] | N | raw pixel | SRC | 6 basic, NE | JAFFE: 213 I | adding noise, black/white mask | noise, block | 95, >91 |
| | [Cotter 2011]* | N | raw pixel | FLSRC | 6 basic, NE | JAFFE: 213 I | black mask, random block | mouth, block | 93, 85 |
| | [Cotter 2010b]* | N | raw pixel | WVSRC | 6 basic, NE | JAFFE: 213 I | random block, black bar | upper/lower face, block | 64, 77, 68 |
| | [Ouyang *et al.* 2013]* | N | LBP map | SRC | 6 basic | CK: 1017 I | black mask, noise | eyes, corruption | 72, 87 |
| | [Zhang *et al.* 2012] | N | raw pixel, Gabor, LBP | SRC | 6 basic, NE | CK: 470 I | noise, replaced block | corruption, block | 68, 42 |
| | [Zhi *et al.* 2011]* | N | GSNMF | nearest neighbor | 6 basic | CK: I | black mask | eyes, nose, mouth | 93, 94, 91 |
| | [Huang *et al.* 2012]* | SRC | STLBP/<br>edge map | WLFF | 6 basic, contempt | CK: 325 I + V | black mask, random mask | eyes, mouth, lower-face, block | 93, 79, 74, 80 |
| | [Liu *et al.* 2014d] | N | raw pixel | MLESR | 6 basic, NE | JAFFE: 213 I<br>CK: 420 I | black mask | random block | 87 (JAFFE), 85 (CK) |
| Sub-region based | [Zhang *et al.* 2014b]*,<br>[Zhang *et al.* 2011] | N | Gabor-based template | SVM | 6 basic, NE | JAFFE: 213 I<br>CK: 1,615 I | white mask | eyes, mouth, random block, clear/solid glasses | 80,78,49,80, 75(JAFFE)<br>95,90,75,95, 92 (CK) |
| | [Song and QiuQi 2011] | N | LBCM | nearest neighbor | 6 basic, NE | JAFFE: I | black mask | eyes, mouth | 94, 93 |
| | [Shuai-Shi *et al.* 2013] | N | LGBPHS | SVM | 6 basic, NE | JAFFE: 213 I<br>natural I | black bar, real-life mask, sunglasses | eyes, mouth, left/right face, scarf, sunglass | 86, 92, 90, 91, 88, 83 |
| | [Liu, et al. 2014c] | N | WLDH | SVM + DF | 6 basic, NE | JAFFE: 213 I | black mask | eyes, mouth, left/right face | 87, 90, 90, 91 |
| | [Lin *et al.* 2013] | GMM | point displacement | EWCCM | 5 AUs, AU comb. | CK+: V | grey block | lower face | 81 |
| | [Dapogny *et al.* 2016] | N | point distance + HOG | WLSRF | 6 basic, NE, 14 | CK+: 1308 I | noisy mask | eyes, mouth | 72,67 (CK+) |
| | | | | | AUs | BU4D: 1212 I | | | 57,49 (BU4D) |
| Statistical model based | [Yongmian and Qiang 2005] | N | point displacement, furrow | DBN | 6 basic, AU | Sample V | missing feature, real-life hand | missing frame, hand | N/A |
| | [Miyakoshi and Kato 2011]* | N | point displacement | BN | 6 basic | JAFFE: 183 I | missing feature | eyes, brow, mouth | 67, 56, 50 |
| | [Tan Dat and Ranganath 2008], [Tan Dat and Surendra 2008] | KLT+ BN | point distance | HMM + ANN | 4 grammatical | CK: V<br>DHHFS: V | real-life hand, blur | hand, blur | 62 |
| | [Hammal *et al.* 2009] | N | point distance | TBM | 6 basic, NE | CFE: 100,800 I | bubble | bubble | N/A |
| 3D data based | [Hu *et al.* 2008] | view classifier | HOG, LBP SIFT + PCA, LDA, LPP | nearest neighbor | 6 basic, 4 intensity | BU3DFE: 12,000 | yaw | (0, 30, 45, 60, 90) | SIFT+LPP:<br>73.1% (avg.)<br>73.9% ($30^o$)<br>71.4% ($90^o$) |
| | [Moore and Bowden 2011] | view classifier | LBP, LGBP, MSLBP, etc. | SVM | 6 basic, 4 intensity | BU3DFE: 48,000<br>Multi-pie: 4,200 | yaw | BU3DFE: (0, 30, 45, 60, 90)<br>Multi-pie: (0, 15, 30, 45, 60, 75, 90) | LGBP: 68% (BU3DFE)<br>81% (Multi-pie) |
| | [Tariq *et al.* 2012] | view classifier | generic sparse coding features | SVM | 6 basic, 4/strongest intensity | BU3DFE: 21,000 | pan;<br>title | (0, ±15, ±30, ±45);<br>(0, ±15, ±30) | 69.1% (4 intensity)<br>76.1% (strongest) |
| | [Vieriu *et al.* 2015] | N | haar on 9 channels | random forest + decision fusion | 6 basic, NE, 4 intensity | BU3DFE: 1400 | yaw;<br>tilt | (-90, 90);<br>(-60, 60) | >66% |
| | [Sun and Yin 2008] | N | shape map | HMM | 6 basic | BU4DFE: 34,200 | yaw;<br>pitch | (-180, 180);<br>(-180, 180) | >80% |
| Deep learning | [Tősér, et al. 2016] | N | similarity normalized images | CNN | 11 AUs | BP4D | yaw;<br>pitch | (-18, 90);<br>(-54, 54) | 55% (F1) |
| | [Cheng, et al. 2014] | N | Gabor | DBM | Six basic, NE | JAFFE: 213 I | black bar | non, eyes, mouth, lower/upper face | 85.7%, 82.9%, 82.9%, 82.9%, 77.1% |
| | [Ranzato *et al.* 2011] | N | raw pixel + MRF + DBN | linear classifier | 6 basic, NE | CK: 327 I<br>TF: 104,000 I | grey block | eyes, mouth, nose, 70% random, right/bottom/top face | N/A |
| | [Batista, et al. 2017] | N | raw pixel | AUMPNet | 7 AUs, 6 intensity | FERA 2017 challenge (BP4D) | pose | 9 different poses | 0.506(F1)<br>0.399(ICC) |
| | [Zhou *et al.* 2017] | view classifier | raw pixel | multi-task deep network | 7 AUs, 6 intensity | FERA 2017 challenge (BP4D) | pose | 9 different poses | 0.879 (RMSE)<br>0.446(ICC) |

Note: "Ocu. Det." stands for occlusion detection. References with * are the top benchmark approaches identified in Section 4.6.
Abbreviations: N – No, T – Texture, G – Geometry, I – Image, V – Video, N/A – Not Available, ANN – Artificial Neural Network, AU – Action Unit, AUMPNet - Unified CNN, BN – Bayesian Network, CFE – California Facial Expression database, DBN – Dynamic Bayesian Network, DF – Decision Fusion, DHHFS - Deaf & Hard-of-Hearing Federation of Singapore, EWCCM – Error Weighted Cross-Correlation Model, HMM – Hidden Markov Model, ICC - Intraclass Correlation Coefficient, KLT – Kanade Lucas Tomasi tracker, KNN – K-Nearest Neighbor, LBCM – Local Binary Covariance Matrices, LGBP - Local Gabor Binary Pattern, LGBPHS – Local Gabor Binary Pattern Histogram Sequence, MLESR – Maximum Likelihood Estimation Sparse Representation, MKL – Mean rule and multiple Kernel Learning, MUG – Multimedia Understanding Group database, MSLBP - Multi-Scale LBP, PCC - Pearson Correlation Coefficient, RMSE - Root Mean Square Error, RW-KNN – Rank Weighted KNN, SD – Salient Detector, SRC – Sparse Representation Classifier, SVM – Support Vector Machine, TBM – Transferable Belief Model, WLDH – Weber Local Descriptor Histogram; WLFF – Weight Learning based Feature Fusion; WLSRF – Weighted Local Subspace Random Forest Model; WVSRC – ρ-Weighted Voting SRC.

Table 3. Main Characteristics of Six Categories of FEA Approaches in the Presence of Occlusion.

| Approach | Face Det. | Ocu. Det. | Face Reg. | Feat. Track. | I/V | Pros | Cons |
|---|---|---|---|---|---|---|---|
| Feature reconstruction | Y | Y(T)<br>N(G) | Y | N(T)<br>Y(G) | I(T)<br>V(G) | robust feature reconstruction based on face configuration. | require reliable feature detector/tracker (G);<br>require precise face alignment/normalization (T);<br>need occlusion detection (T);<br>loss of texture. |
| Sparse representation | Y | Y/N | Y | N | I | optimal feature representation;<br>estimate occlusion location. | require precise face alignment/normalization;<br>assume test and training data are linearly correlated. |
| Sub-region based | Y | N | Y | N | I | easy to implement;<br>good result for small occlusion;<br>not need occlusion detection. | require precise face alignment/normalization;<br>dependent on face subdivision;<br>dependent on decision fusion. |
| Statistical model based | Y | N | N | Y | I/V | robust via temporal reasoning;<br>close to real situation. | require robust feature trackers;<br>difficult to create ground data. |
| 3D data based | Y | Y/N | Y | Y | I/V | depth information;<br>robust to pose variations. | require face view classifier & view-dependent emotion classifier;<br>require mapping to frontal view;<br>heavy computation. |
| Deep learning | Y | N | N | Y | I/V | Automatic feature extraction;<br>no need occlusion detection. | require large training data;<br>fine-tuning large system parameters;<br>heavy computation. |

Note: "Face Det.", "Ocu. Det.", "Face Reg.", "Feat. Track." stand for face detection, occlusion detection, face registration and feature tracking respectively.

Abbreviations: Y – Yes, N – No, T – Texture, G – Geometry, I – Image, V – Video. Take the step of occlusion detection for instance, 'N(G)' means that occlusion detection has not been adopted for constructing geometric features in existing studies, and 'Y/N' means that occlusion detection have been used in some existing studies, while not used in other studies.

## 4.1 Feature Reconstruction Approach

The feature reconstruction approaches attempt to overcome the effect of occlusion by reconstructing missing geometric and (or) texture features caused by partial occlusion based on the visual configuration of the face, and they are the most popular approach in early FEA studies on handling occlusion. Approaches in this category can be further grouped into *geometry based* and *texture based*, according to the type of the features used for representing emotions.

### 4.1.1 Geometry based Approach

One group of geometry based approaches focuses on utilizing coordinates of facial points in static images. Towner and Slater [2007] compared the performance of three PCA based methods in the reconstruction of the positions of missing feature points at the top and bottom parts of the face. They found that the conditional mean method produced the best accuracy for recovering random subsets of 22 points on the CK database. The feature points reconstructed from partially occluded images were then used as the input of a Support Vector Machine (SVM) for classifying six basic emotions. There is only a 5% and a 3% reduction in overall classification rates for the top and bottom occlusion respectively, compared to using all facial points. Zhang *et al.* [2015] combined Iterative Closest Point (ICP) features and the Fuzzy C-Means (FCM) algorithm to reconstruct 54 facial points in an occluded face using prior knowledge of facial elements. The geometry of facial points was used to predict 18 AUs and eight emotions using Support Vector Regression (SVR) and Artificial Neural Networks (ANNs). The experiments showed more than 78% point detection accuracy under occlusion of the top and bottom parts of the face. However, the performance for facial expression recognition was not evaluated.

The other group is based on temporal geometric coordinates of facial points in video sequences. Bourel *et al.* [2001] proposed using recovered geometric features to handle occlusion in regions of the upper face, mouth, and left/right half of the face in video sequence (Fig. 4). Their approach adopted an enhanced version of the Kanade-Lucas tracker to reconstruct drifting or lost facial points during face tracking, and then generated local spatiotemporal vectors based on geometrical positions of 12 facial points. A rank-weighted K-Nearest Neighbor (KNN) classifier was further applied independently to local facial regions to ensure that the occluded region does not affect other regions. A *sum* fusion of classifier outputs was finally used to obtain an emotion label for each sequence. The method achieved more than 80% accuracy for classifying four emotions under four types of occlusion using 100 CK sequences, except for sadness under an occluded mouth with around 20% accuracy. However, the spatiotemporal vectors in [Bourel 2001] suffer from a lot of noise. To reduce the noise, Bourel *et al.* [2002] further converted continuous values into three discrete states (*increase, stable* and *decrease*) based on the average motion amplitude of a sequence representing a facial expression. It resulted in higher accuracy for classifying six basic emotions under the four types of occlusion plus random noise using 300 CK video sequences. However, both approaches require manual annotation of facial points in the first frames of video sequences, and need to manually omit the occluded features from feature vectors.

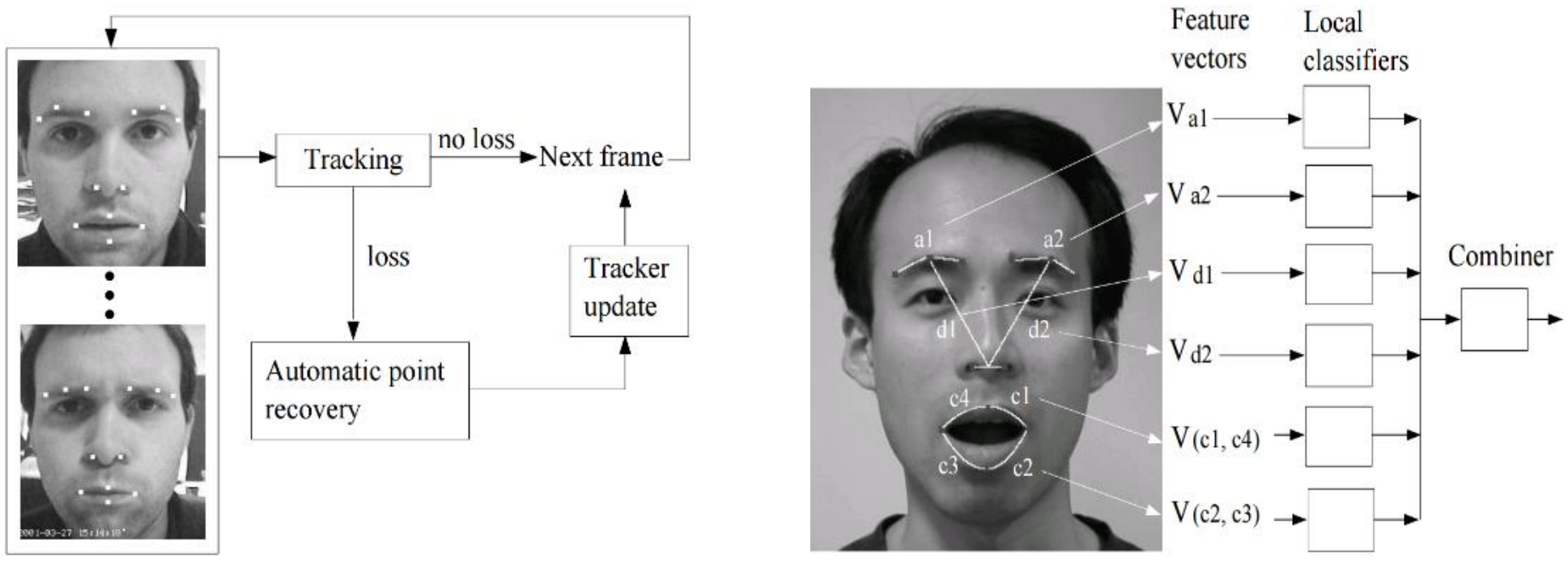

**Fig. 4. (left) Recovery of lost or missed geometric facial landmarks using an enhanced Kanade-Lucas tracker in video sequences. (right) Multiple local classifiers are generated from geometric features and fused to derive a class label [Bourel 2001].**

Different from the above approaches which often depend on data from normal digital camera, Kapoor *et al.* [2003] proposed an AU recognition system based an infrared sensitive camera. The camera was employed to robustly detect the pupil location even under unfavorably lighting conditions, and the pupil location was used to find and normalize the eyes and brow regions. The x and y coordinates of landmark points from the eyes and brows were then extracted as shape parameters. The parameters in occluded regions in test images were recovered by finding a linear combination of example images using Principal Component Analysis (PCA). 69.3% accuracy was achieved for each individual AU classification in real-life video frames with head motions, occlusion and pose changes. Details about occlusion, however, were not given by the authors.

#### 4.1.2 Texture based Approach

Existing texture based approaches are mainly based on the Robust PCA (RPCA) algorithm. Xia *et al.* [2009] combined RPCA and saliency detection for FEA under occlusion. The occlusion was located using a saliency detector by setting a threshold to the pixel difference between the occluded image and the reconstructed image using RPCA. The occluded region was then replaced by the corresponding region in the reconstructed face. Haar-like features were further extracted and fed into AdaBoost for classifying six basic emotions. The method produced 5% and 16% higher accuracies for hands and sunglasses occlusion respectively than using Adaboost alone on the JAFFE and BHUFE databases. Rather than using Haar-like features, Cornejo *et al.* [2015] extracted Gabor wavelets and a geometric representation of 22 facial points from the recovered regions of occlusion using RPCA. A KNN or SVM classifier was adopted for recognizing expressions in the presence of five types of occlusion, including two eyes, left eye, right eye, bottom left or bottom right part of the face, and bottom part of the face. The approach achieved more than 98% accuracy for random partial occlusion on the JAFFE, CK+ and MUG databases. In another similar work by Cornejo *et al.* [2016], the CENsus Transform hISTogram (CENTRIST) features were extracted from RPCA reconstructed facial regions, and further fed into PCA plus LDA for feature reduction, and finally SVM for emotion recognition. The approach showed 92% and 90% accuracies on the JAFFE and CK+ datasets. The RPCA was also found to outperform PCA and perform similarly to probabilistic PCA in reconstructing occluded eyes and mouth [Jiang and Jia 2011].

The feature reconstruction approaches retain promising robustness against occlusion by recovering the features in occluded regions based on facial configuration characteristics. However, they are still heavily dependent on reliable face detection and facial feature tracking. Approaches used for reconstructing geometric features face the challenge of accurately detecting and robustly tracking geometric points in the presence of partial occlusion. Approaches [Bourel 2001], [Bourel, et al. 2002] which depend on manual annotation of the coordinates of facial points in the first video frames have problems in automatic recognition. Approaches to reconstructing texture require pre-locating occluded regions and precise face alignment, which are still challenging issues. PCA based techniques have difficulty of precisely recovering dynamically varied local and subtle texture in an unseen facial region based on information learnt from the training data, which may significantly impact the performance. In brief, current feature reconstruction approaches are heavily dependent on the accuracy of face detection and tracking. As a result, it is still difficult to achieve fully automatic implementation of these approaches in real-life applications.

### 4.2 Sparse Representation Approach

The sparse representation approach was firstly proposed for face recognition tasks in [Wright *et al.* 2009], and later applied into FEA, especially from occluded faces [Cotter 2010a],[Cotter 2010b],[Cotter 2011],[Zhang, et al. 2012]. The approach treats all training samples as a dictionary and performs robust object recognition using a sparse representation of a test image that is formed by finding a linear combination of training images from the same class. The optimal weights for combining training images are searched through solving a convex optimization problem via $l_1$ minimization. For the purpose of handling occlusion, the error caused by occlusion is represented by an individual identity matrix that can be isolated from the feature matrix of un-occluded training images. For occlusion that is not overly large, the error matrix can be calculated using $l_1$ minimization. A clear image can be then recovered by subtracting the calculated sparse solution of the error matrix from the test image at the emotion classification stage.

Cotter [2010a] is one of the first work in applying the Sparse Representation Classifier (SRC) into FEA in noise corrupted or occluded JAFFE images. The SRC obtains 95% accuracy when 50% pixels are corrupted by noise and over 91% accuracy for block occlusion with a size ranging from 10×10 to 40×40 pixels. The SRC outperformed Eigenfaces or Gabor features with an ANN or SVM classifier on average. It was found that the use of a black or a white color in simulating block occlusion also impacts the performance. To more effectively utilize the local characteristics of facial occlusion, the SRC was later extended to perform on sub-regions of the face. Cotter [2011] proposed the Fusion of Local SRC (FLSRC), which performs SRC separately in each of three facial regions – the mouth, left eye and right eye. Unoccluded regions are dynamically determined by setting a threshold to representation errors in SRC. Features from only unoccluded regions are used to select the most important individual region (or fused region) to make a final classification decision. The tests using JAFFE images with an occluded mouth and randomly placed block occlusion showed that fusion of all regions leads to higher accuracy than using each alone. The FLSRC significantly outperformed SRC for an occluded mouth (93.4% vs. 72.8%). For small random block occlusion, they performed similarly, but FLSRC was superior for block occlusion larger than 35×35 pixels in 96×72 facial regions. Unlike the FLSRC method in [Cotter 2011], which selects only the most important facial region in making a decision, a ρ-Weighted Voting SRC (WVSRC) scheme was used in [Cotter 2010b]. The scheme assigns different weights to decisions from nine equally divided facial sub-regions and then combines weighted decisions. The weights are assigned based on the class representation error of SRC in each sub-region. The WVSRC significantly outperformed both SRC and Gabor based approaches for occluded upper and lower halves of the face, and large randomly placed block occlusion. However, a direct comparison between WVSRC and FLSRC was not reported.

The original sparse representation assumes a Gaussian distribution of the coding residual, which may suffer from inaccuracy in describing errors in practice. Liu *et al.* [2014d] proposed the Maximum Likelihood Estimation Sparse Representation (MLESR), which models the sparse coding task as a sparsely constrained regress problem and iteratively assigns lower weights to pixels in occluded regions until the result converges. A test image was classified into an emotion which has the minimal spare representation residual with this image. The MLESR outperformed SRC and Gabor based SRC (GSRC) for all degrees of simulated random occlusion ranging from 0 to 90% of image pixels on the JAFFE and CK databases. Zhi *et al.* [2011] presented a Graph-preserving Sparse Non-negative Matrix Factorization (GSNMF) algorithm to utilize both the sparse and graph-preserving properties of facial images. The GSNMF transforms a high-dimensional image into a low-dimensional locality-preserving subspace to achieve robustness to partial occlusion. The GSNMF with the nearest neighbor classifier achieved 93.3%, 94.0% and 91.4% accuracies for classifying six basic emotions under the eyes, nose and mouth occlusion respectively on the CK database.

Unlike the above work that applied SRC directly on raw image pixels, studies also exploited other feature descriptors. Ouyang *et al.* [2013] suggested using Local Binary Pattern (LBP) maps that generally have good robustness against illumination variations. Fusion of LBP maps and SRC was found to outperform SRC under non-occlusion, occluded eyes, and partial corruption conditions on the CK database. The fusion achieved 87.0% accuracy when 35% of the face is replaced with uniform distributed values and 72.4% accuracy when 30% of the face is occluded in the eyes region. Zhang *et al.* [2012] performed performance comparisons among raw pixels, Gabor wavelets and LBP features, as well as SRC, ANN, SVM and the nearest subspace classifiers. Results on CK images indicated that Gabor features with SRC performed the best for non-occluded images. SRC significantly outperformed ANN, SVM and the nearest subspace at various levels of random pixel corruption and random block occlusion. Huang *et al.* [2012] investigated the use of Spatio-Temporal LBP (STLBP) and edge map features for FEA under occlusion in video sequences (Fig. 5). The features were extracted from the mouth, nose and eyes components and further integrated in sparse representation to generate a binary codebook for determining occluded components. For expression recognition, the features were also combined via a feature-level fusion and their optimal weights were learned using multiple kernel learning. The system yielded 93%, 79.1% and 73.5% accuracies for classifying six basic emotions plus contempt under occlusion of the eyes, mouth and lower-face respectively in CK+ frames. The integration of occlusion detection led to higher accuracies for most emotions. A recent study [Amirian *et al.* 2017] using sparse coding also achieved promising results of AU intensity estimation in the FERA 2017 challenge. The proposed approach first estimated the head pose using dictionary learning and then computed a sparse representation of image patches to train a Support Vector Regression for AU intensity estimation. The approach produced Intraclass Correlation Coefficient (ICC) of 0.295 and Root Mean Square Error (RMSE) of 0.970 on the test subset of the challenge.

To summarize, the great advantage of the sparse representation approaches is that they are not only robust to small occlusion and corruption, but also can be used to estimate the occluded or corrupted parts of the face. The approaches have been demonstrated as one of the most promising techniques in overcoming occlusion for FEA. However, their performance is largely dependent on whether the test data can be accurately represented using a linear combination of a subset of training samples from the same emotion, and the availability of a reasonable large number of training samples with sufficient variations for the emotion. The training dictionary needs not only sufficient information to effectively represent the test data, but also abundant characteristics to reduce the correlations of training samples from different classes [Ouyang, et al. 2013]. To ensure accurate feature extraction, the approaches require precise face location, alignment and normalization, which are done primarily manually in existing work [Cotter 2010a],[Cotter 2010b],[Zhang, et al. 2012],[Ouyang, et al. 2013]. One important factor in the use of SRC is to choose a proper type of feature descriptor. Although several feature descriptors such as raw pixels, Gabor and LBP, have been investigated in existing studies, it is still a largely unexplored field regarding which descriptor works best for handling facial occlusion. It is still worth investigating other types of descriptors such as Scale-Invariant Feature Transform (SIFT) and Histogram of Oriented Gradients (HOG) to further improve the performance.

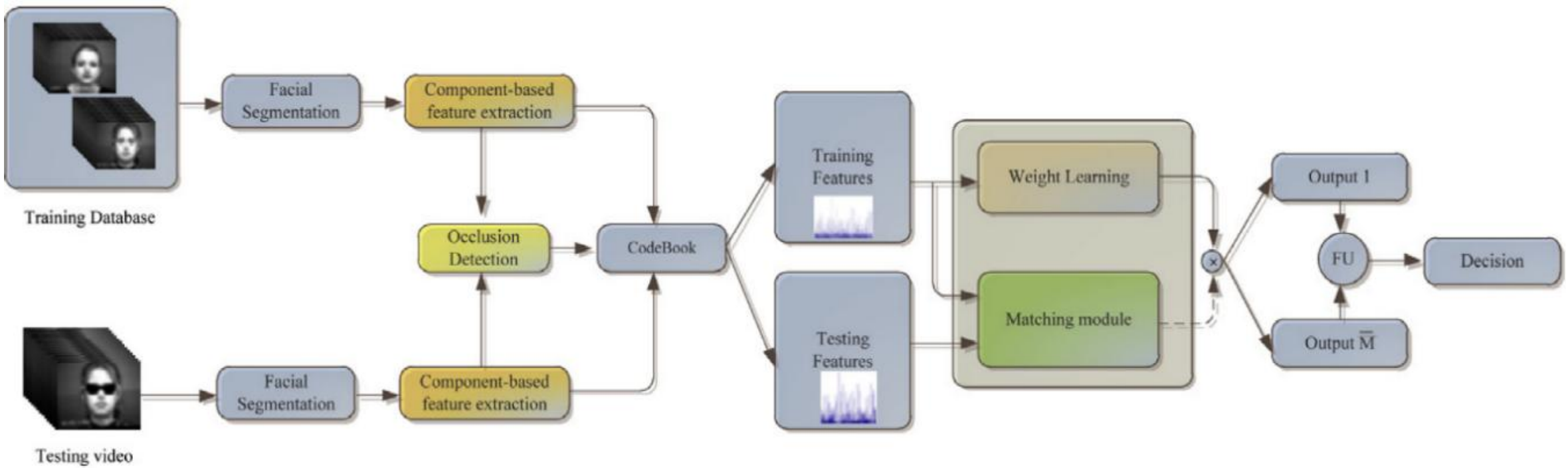

**Fig. 5. An automatic system comprising of occlusion detection using STLBP based sparse representation, and emotion recognition via multiple feature fusion with weights obtained using multiple kerning learning [Huang, et al. 2012].**

## 4.3 Sub-region based Approach

The sub-region based approaches treat the whole facial region as comprising of a set of local sub-regions and attempt to fuse information from only non-occluded sub-regions for FEA. The approaches can be approximately categorized into 1) *feature fusion approach* which fuses features from sub-regions, and approaches in this category often adopt a feature selection mechanism to remove the features extracted from occluded sub-regions, and 2) *decision fusion approach* which fuses classification decisions from sub-regions, and approaches in this category often employ a voting strategy to reduce the weights of decisions derived from occluded sub-regions.

### 4.3.1 Feature Fusion Approach

Zhang *et al.* [2011],[2014b] proposed Gabor based templates for FEA under the eyes, mouth, glasses and randomly placed block occlusion. A Monte Carlo algorithm was used to collect a group of multi-scale 3D Gabor templates from randomly selected locations in the gallery images. The collected templates form a pool of local texture features, and thus it is anticipated that only a small portion of them are significantly impacted by partial occlusion. A template-based matching process was then performed over a local search area to create distance features which encode high-level expression information of the face and suffer from limited impact by occlusion. A linear SVM was further used to select a small set of most effective distance features from the whole feature set. The SVM employs normal based feature selection [Dunja *et al.* 2004] by treating the normal to the classification hyperplane as weights of features and keeping only those features with high weights. Those features often have the biggest impact on the classification results and thus are most important. The six basic emotions plus neutral were finally classified using another linear SVM classifier. The experiments showed that a larger occlusion has bigger influence on the overall performance of both JAFFE and CK databases. The eyes region and the mouth region have big effects on JAFFE and CK, respectively. The system showed robustness to changes in parameters of Gabor filters and template sizes. The work [Zhang, et al. 2011],[Zhang, et al. 2014b] adopted features extracted from a large number of random local patches to overcome facial occlusion, while another commonly used way is to extract features from a set of facial sub-regions. Guo and Ruan [2011] used a *sum* function to combine Local Binary Covariance Matrices (LBCM) features from nine equally sized facial sub-regions. The robustness to occlusion was achieved by removing the sub-region with the maximal distance between covariance matrices from the gallery and the probe sets. Using the nearest neighbor classifier, the LBCM outperformed Gabor filters and SRC, and achieved 93.7% and 92.9% accuracies for occluded eyes and mouth respectively using JAFFE images. Liu *et al.* [2013] investigated a feature-level fusion of Local Gabor Binary Pattern (LGBP) maps from equally divided facial sub-regions. Tests on occlusion of the mouth, eyes, left and right face gave 92.1%, 85.5%, 89.5% and 90.8% accuracies respectively on the JAFFE database. An additional test on 24 natural images from a person with real-life medical mask and sunglasses occlusion showed 87.5% and 83.3% accuracies for classifying four expressions.

### 4.3.2 Decision Fusion Approach

Liu *et al.* [2014c] employed a *maximum* decision fusion of equally sized facial sub-regions (Fig. 6). From each sub-region, Weber Local Descriptor Histogram (WLDH) features were extracted and fed into an SVM classifier. The SVM outputs from all sub-regions were fused using a *maximum* function. The approach achieved more than 87% accuracy for occlusion of the mouth, eyes, left and right sides of the face using JAFFE images. Dapogny *et al.* [2016] presented Local Expression Predictions (LEPs) for categorical FER and AU prediction under partial occlusions. The LEPs were generated by locally averaging predictions by local trees in random forests which are trained using random facial masks generated in specific parts of the face. HOG features extracted from facial landmarks were used as the local descriptor of facial masks. For occlusion-robust FER, local confidence measurements were obtained based on the reconstruction error outputted by a two-layer autoencoder network to weight LEPs in different facial parts. The network was trained to model the local manifold of non-occluded faces and to reconstruct feature patterns in an occluded face, providing a confidence measurement based on the reconstruction error. Occlusions were simulated by overlaying noisy patterns to regions of eyes and the mouth in facial images. Evaluations on the CK+ and BU4D databases showed around 72% and 57% under occluded eyes, and 67% and 49% under occluded mouth for categorical FER. Instead of using texture features, Lin *et al.* [2013] predicted AUs under mouth occlusion using geometric features of displacements of facial points. A Gaussian Mixture Model (GMM) was employed to model the gray pixel distribution in facial regions for detecting the occluded region. Facial Deformation Parameters (FDPs) were represented using the displacements of 74 landmark points in six regions, including the mouth, eyes, eyebrows, nose, cheeks, and jaw. The FDPs in each region and the relationships among paired regions were modelled using a Cross-Correlation Model (CCM), and the prediction decisions of CCM in all paired regions were finally combined using a Bayesian weighting scheme. 80.7% accuracy was obtained in predicting five AUs or AU combinations in CK+ sequences. To optimize the fusion of decisions from nine equally divided facial sub-regions, a $\rho$-weighted voting SRC scheme was presented in [Cotter 2010b] to assign a different weight to each sub-region.

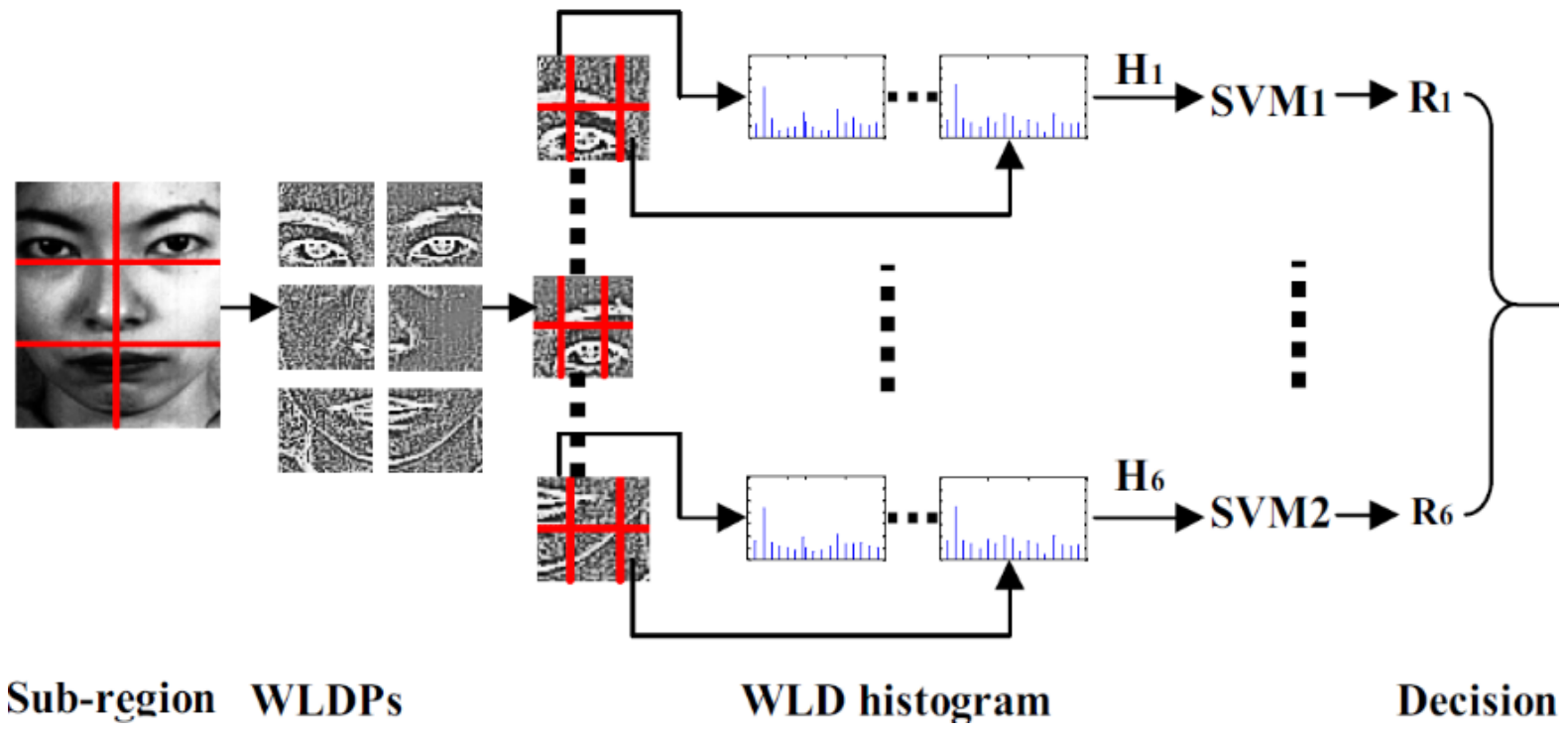

**Fig. 6. WLDP features and SVM classifier are used to derive a classification decision for each facial sub-region. The expression classification is achieved using a maximum decision fusion of all sub-regions [Liu, et al. 2014c].**

A basic assumption of sub-region based approaches is that occlusion presents only in a small portion of the face, and thus its effect can be minimized via a feature selection method or a decision voting strategy over all sub-regions. The approaches are often capable of producing satisfactory performance for small occlusion. However, the granularity of subdividing the face into local regions, and its effect on the performance is still an open issue, particularly for random occlusion without a fixed location, shape and size. Similar to feature reconstruction and sparse representation approaches, the sub-region based approaches are also sensitive to noise due to inaccurate face location, alignment and normalization. It is still an issue to choose a proper fusion strategy for multiple results [Bourel 2001], such as a linear combination, fuzzy logic, and ANN. State-of-the-art sub-region based approaches are still at the beginning stage of exploiting the most effective facial region division method and the best decision or feature fusion strategy.

## 4.4 Statistical Model based Approach

Unlike the above approaches, the statistical model based approaches do not directly reconstruct the features in occluded regions nor divide face into sub-regions. Instead, they try to infer occluded features using statistical prediction models by utilizing the temporal correlations between neighboring video frames or spatial dependent information in non-occluded parts in static images. A unique feature of statistical model approaches lies in the capacity of robustly inferring facial features in a current frame based on facial information in neighboring frames, even when the current frame contains a partially occluded face or is a completely missing frame.

Hammal *et al.* [2009] exploited the usefulness of facial point deformations with a modified Transferable Belief Model (TBM) for recognizing facial expressions from images with partial occlusion. Five distances from the contours of the mouth, eyes, and eyebrows were normalized and mapped to symbolic states. The TBM has the advantage of automatically integrating features from multiple local facial regions and handle uncertain or imprecise data such as occluded facial regions, and thus it is adopted for modeling the correlation between expressions and symbolic states. The results on 70 facial images with occlusion simulated by *bubble* masks showed that the use of all five distances obtained the highest rates for recognizing happiness (100%), anger (100%), surprise (75%) and disgust (75%), but a much lower rate for recognizing sadness (25%). The behaviors of human observers are different from TBM based models and the human tends to use "suboptimal" features for FEA under occlusion. Rather than using distances between facial components in static images [Hammal, et al. 2009], Miyakoshi and Kato [2011] improved this by using movement magnitudes of 14 points from a neutral face to an emotional face in static images. A Bayesian network classifier was employed to learn the dependencies between target facial expressions and facial features without involving a process of filling in the facial gap due to occlusion. The causal relationships among facial features and the structural associations between expressions and facial features were learnt using the $K_2$ algorithm and stepwise feature selection respectively. 67.1%, 56.0%, and 49.5% accuracies were observed for occlusion of the eyes, brows and mouth in JAFFE images.

Given the limited information available in static images, other statistical model based approaches have focused on directly processing video sequences. Work [Tan Dat and Ranganath 2008],[Tan Dat and Surendra 2008] designed a Bayesian tracker to reliably track facial points in the presence of temporal occlusion by head motions or hands (Fig. 7). The tracker augments the Kanade Lucas Tomasi (KLT) tracker by additionally incorporating a Bayesian feedback mechanism. Seven eyebrow and four eye distances were extracted from the tracked points and fed into seven individual Hidden Markov Models (HMMs) for classifying four face movement categories and three head motions. The likelihood outputs of HMMs were further used as inputs into an ANN for classifying four grammatical expressions (Yes/no, Wh, Topic and Negation). The proposed tracker showed more stable tracking results than KLT. The proposed system had an accuracy drop from 69% to 62% in the presence of hand occlusion and performed the best when it was trained and tested using the same tracker. However, the system requires manual annotation of facial features in the first video frames. Unlike [Tan Dat and Ranganath 2008],[Tan Dat and Surendra 2008], which aimed to generate robust facial point trackers, Yongmian and Qiang [2005] focused on recognizing AUs directly from occluded video frames (Fig. 8). They utilized temporal reasoning via Dynamic Bayesian Networks (DBNs), which are capable of accounting for information from both current visual observations and previous visual evidences. Facial features missed in an occluded frame were compensated by modeling its temporal correlations with neighboring frames. Although no systematic evaluation was conducted, impressive results were achieved using video examples with undetected or untracked frames and hand occlusion.

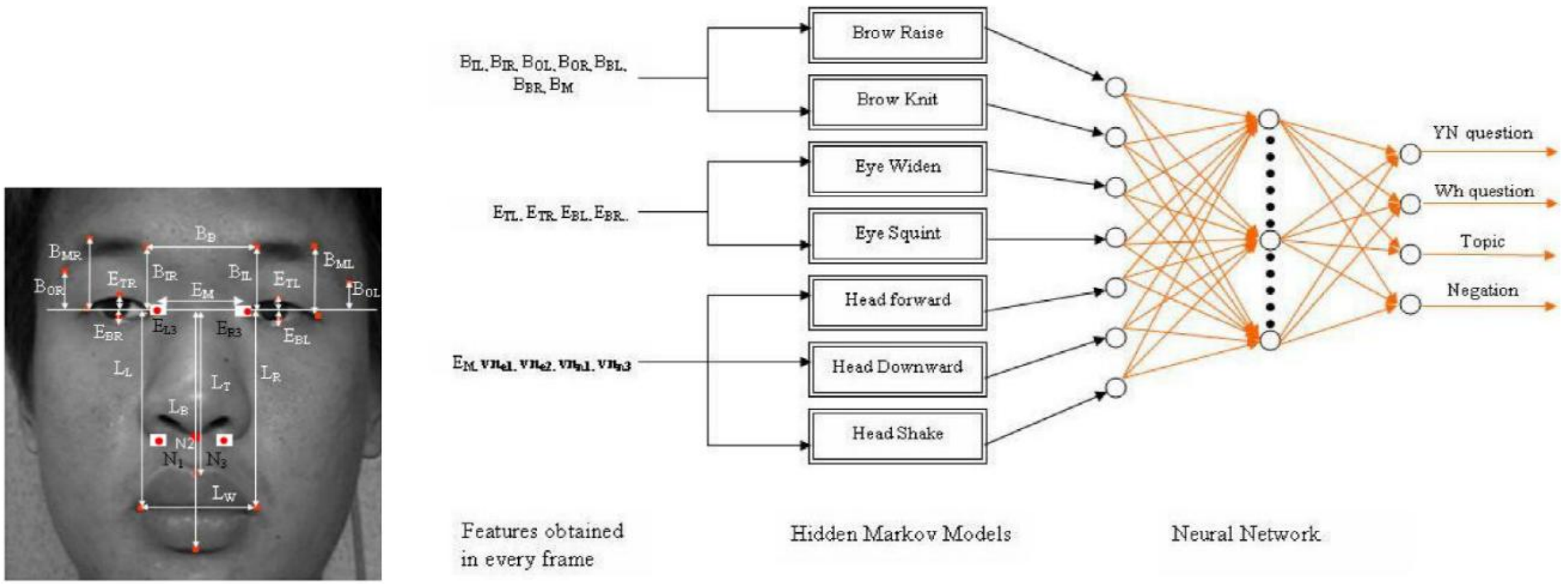

**Fig. 7. Distance features between facial points are extracted based on a Bayesian tracker, and they are fed into individual HMMs and ANNs subsequently for classifying four grammatical expressions [Tan Dat and Ranganath 2008].**

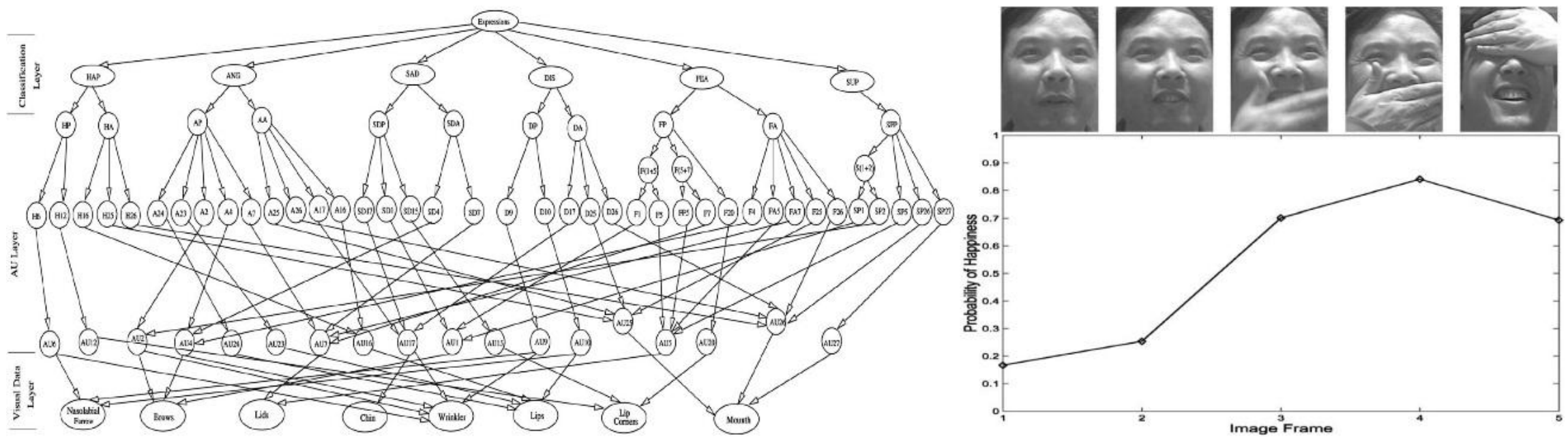

**Fig. 8. (left) Dynamic Bayesian networks generated for FEA and (right) its accuracy for classifying happiness under temporal occlusion caused by a moving hand [Yongmian and Qiang 2005].**

Image based statistical model approaches generally depend on spatial relationships in facial parts learnt from training data to recover facial occlusions, but they also require large training data, and heavy computation. On the other hand, the greatest advantage of video based statistical model approaches is that they are able to utilize temporal information in video sequences to infer features in occluded facial regions, and thus the results are anticipated to be more robust than static image based approaches because facial expressions often exhibit strong and unique temporal patterns and correlations. This approach also shows a good capacity of handling missed frames and is closer to the situation of handling facial occlusion in real-life applications. The drawbacks lie in the difficulty of creating suitable ground truth video data that represents a full sequence of a facial expression to train the model, and the requirement of robust facial feature trackers in occluded frames. To conclude, statistical model based approaches seem to be a robust method for handling facial occlusion during FEA. However, due to the lack of suitable benchmark databases, direct comparisons of statistical model based approaches with other approaches are still largely unexplored in existing work. As a result, their performance advantages over other approaches still need to be further validated.

## 4.5 3D Data based Approach

Most approaches discussed so far are based on 2D facial data. The 3D data based approaches include the additional depth information about the facial structure and appearance on top of the normal 2D data. The depth information can be potentially utilized for generating more robust, discriminative, and view-independent features under facial occlusion, particularly those caused by head pose changes, or missing parts [Drira *et al.* 2013]. Although FEA using 3D data has been researched intensively (see [Sandbach, et al. 2012] for a recent survey on existing studies), the approaches that are specifically designed to overcome facial occlusion are still limited and largely focus on handling self-occlusion caused by head pose variations. This part introduces several typical 3D based approaches for FEA with or without facial occlusion.

One typical 3D data based approach is *multiple views* method, which first estimates the face's current view angle and then builds a separate emotion classifier for each angle. Accordingly, there are generally two sequential steps, including view classification and view dependent FER. Hu *et al.* [2008] adopted a five-class view classifier to determine the view and trained a separate emotion classifier for each view. Three descriptors - HOG, LBP and SIFT, and three feature dimension reductions – PCA, LDA and Locality Preserving Projection (LPP), are comparatively used with the same nearest neighbor. Experiments were conducted on recognizing 6 emotions from 12,000 face data with five yaw angles (0, 30, 45, 60 and 90 degrees) on the BU-3DFE database. The results showed a fusion of SIFT and LPP produced the lowest average error rate of 26.9% for all views. Moore and Bowden [2011] compared LBP features with their variations as texture descriptors for both facial view and expression classification using a multi-class SVM. Face images with five yaw angles (0, 30, 45, 60 and 90 degrees) are projected from a 3D textured model on the BU-3DFE database. Experiments indicated that Local Gabor Binary Patterns (LGBPs) perform the best, particularly at large yaw angles. Tariq et al. [2012] utilized generic sparse coding features with a linear SVM for multi-view FER. Experiments used 84,000 face data with seven pan angles (0, ±15, ±30 and ±45 degrees) and five title angles (0, ±15 and ±30 degrees) from the BU-3DFE database. The approach achieved 69.1% and 76.1% accuracies of classifying six emotions with four emotion intensities and only the strongest intensity, respectively.

Another type of 3D-based approach is *simulated 3D features* method. Different from *multiple views* methods, this method directly performs FER on 3D non-frontal facial data by feature mapping. The mapping is achieved by transferring non-frontal features to their counterparts in a frontal view of the same face. Vieriu *et al.* [2015] transformed 3D data of the face onto a pose invariant 2D cylindrical representation (Fig. 9), where self-occlusion was treated as missing information in this representation. The representation was later split into multiple overlapping patches, and from each patch, Haar features were extracted from 9 channel maps and a local random forests classifier was generated. The emotional state of the face was recognized via a weighting decision scheme which fuses probabilities from patch-specific random forests. Evaluations showed that the method achieved a recognition rate of 66.2% on faces with (-90, 90) degrees of yaw rotation and (-60, 60) degrees of tilt rotation on the BU-3DFE database. Rudovic *et al.* [2010] mapped locations of 39 points in a non-frontal face to their corresponding locations in a frontal view using four regression functions, which were fed into a frontal face SVM emotion classifier. Evaluations on classifying four emotions from 800 facial images with four views (0, 15, 30 and 45 degrees) on the Multi-PIE database showed that the approach outperformed view-specific classifiers.

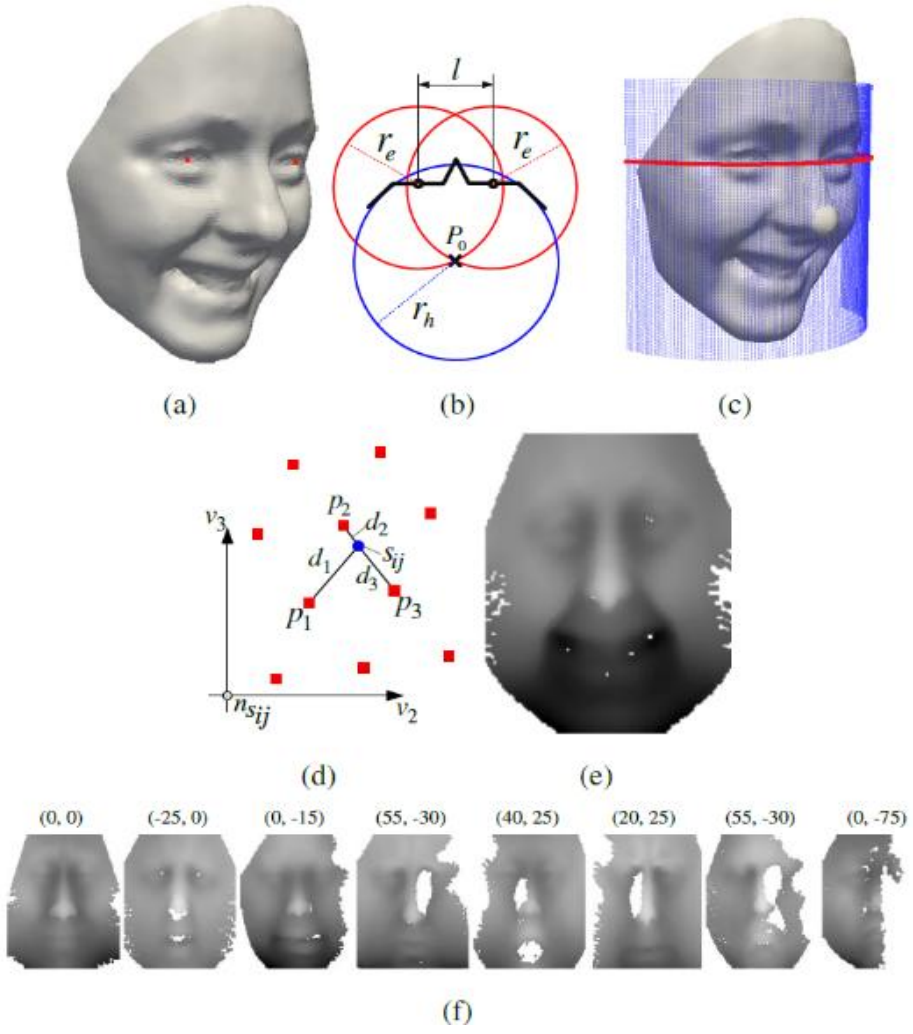

**Fig. 9. Projection of sampled 3D faces under varying head poses in (f) to a frontal pose-invariant face representation in (e) using a cylindrical head model in (b) [Vieriu, et al. 2015].**

There are also other 3D-based methods that are not specifically designed to handling facial occlusion, but can be potentially used for FEA under occlusion. Sun and Yin [2008] presented a spatio-temporal 3D model-based approach which integrates 3D facial surface descriptor and HMM for emotion recognition. The surface descriptor included eight primitive shape types and was generated based on 83 facial landmarks and LDA based feature reduction. The approach was evaluated on simulated partial occlusion by changing the yaw and pitch angles of a 3D face model. The results showed over 80% accuracy for raw and pitch angles of 60 degrees, and only limited decreased accuracy for a yaw of close to 90 degree. However, the accuracy degrades to zero when either pitch or yaw changes to a 150 degree where useful facial information is completely occluded. The results also indicated that temporal approaches outperformed image based statistic approaches, implying that motion information helps to compensate the loss of spatial information. In [Zhao *et al.* 2011], a 3D Statistical Facial feAture Model (SFAM) was presented to locating facial landmarks in the presence of facial expressions and occlusion. The SFAM combined global variations in 3D face morphology and local variations of texture and geometry features around each facial landmark, and integrated them in an objective optimization function. To identity the type of occlusion, a histogram of the similarity map between local shapes of the target face and shape instance from the SFAM was used with a KNN classifier. The type of occlusion was also integrated in the optimization function to localize facial landmarks by setting a binary weight (0 for occluded and 1 for unoccluded) to each landmark region. Experiments on the Bosphorus database demonstrated an accuracy of 93.8% for classifying four types of occlusion (i.e., eyes, mouth, glass and non-occlusion) and a precision of 20-mm for locating 97% landmarks. The SFAM also achieved a precision of less than 10mm for most landmarks on the FRGC V1, V2, and BU-3DFE databases. However, the SFAM was designed for landmark detection and no result on FER was reported.

Studies also investigated fusion of 2D with 3D feature descriptors to improve FER results. Li *et al.* [2015] represented the local texture around 2D facial landmarks using histogram of second order gradients and first-order gradient based SIFT descriptor, and the local geometry around 3D facial landmarks using histogram of mesh gradients and histogram of mesh shape index. The texture and geometry were fused at both feature-level and score-level to improve FER based on an SVM classifier. Evaluations on the BU-3DFE and Bosphorus databases show that 2D and 3D descriptors provide complementary characteristics (Fig. 10). Zhao *et al.* [2016] localized facial landmarks from 2.5D facial data using a deformable partial face model. Global and local features were extracted from those landmarks and used to represent coordinates in an AU space, where each region was classified to a specific affective state using SVM. The approach achieved promising FEA results on the EURECOM, FRGC and Bosphorus databases. However, those studies were not designed for handling facial occlusion.

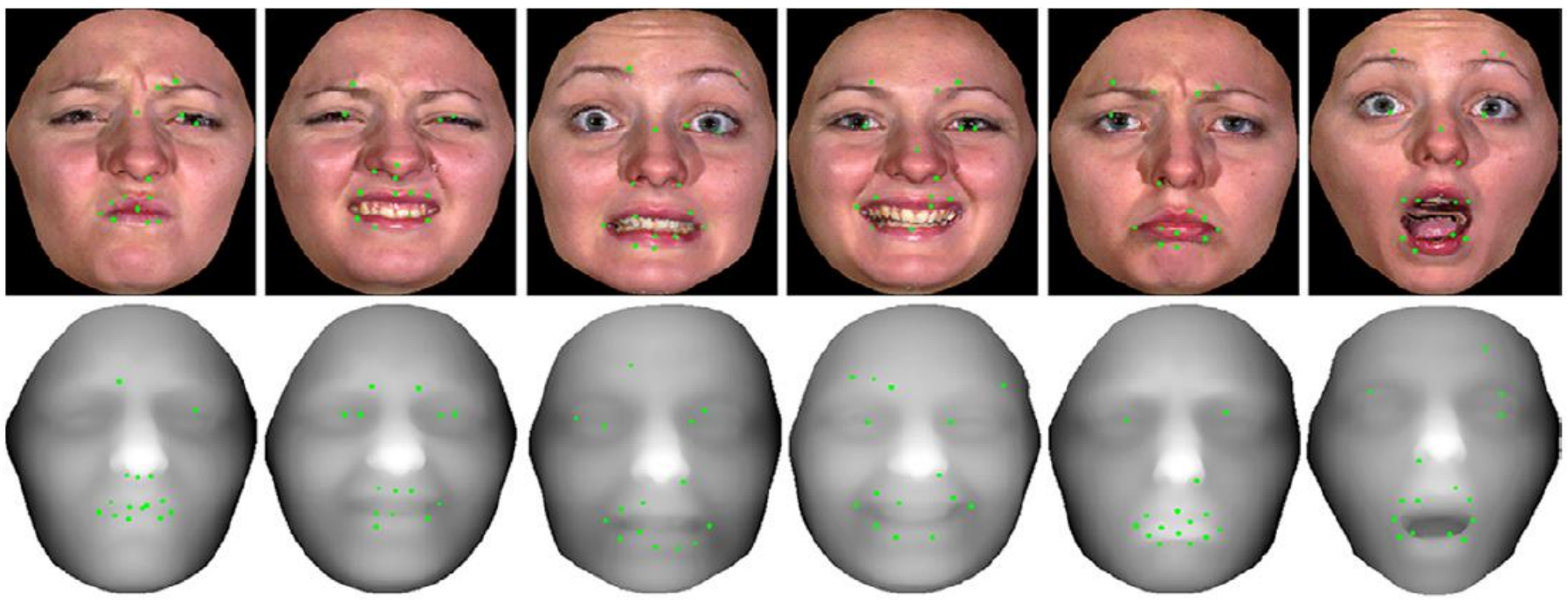

**Fig. 10. Distribution of top 15 most discriminative 2D (top) and 3D (bottom) landmarks. The 2D and 3D features provide complementary characteristics for FEA [Li, et al. 2015].**

A benefit of using 3D data is the possibility of using a richer set of information about facial structure, texture, and depth. Thus, 3D data based approaches are more robust in handling facial occlusion, particularly self-occlusion caused by head pose variations. However, the disadvantages are heavy computational time required and the necessity of designing algorithms to accurately map non-frontal facial features into a frontal face representation. Existing studies have been heavily dependent on several available databases collected in controlled laboratory environments. The lack of naturalistic 3D FEA databases with different types of occlusion has largely restricted the current progress on designing 3D data based approaches and evaluating their performance in realistic conditions.

## 4.6 Deep Learning Approach

In recent years, deep learning approaches such as Boosted Deep Belief Network [Liu *et al.* 2014b], Convolutional Neural Networks (CNNs) [Chang *et al.* 2017], Long Short-Term Memory [Rodriguez *et al.* 2017], and fusion of multiple deep neural networks [Zhang *et al.* 2017], have gained increasing popularity due to their ability to deliver state-of-the-art performance on FEA (see a recent review in [Zafeiriou, et al. 2016]). Deep learning has advantages of learning more abstract patterns progressively and automatically from raw image pixels in a multiple layer architecture rather than using hand-engineered features. It is better suited to learn embedded patterns in facial regions and encode geometric relationships between facial components, and thus it can be used in recovering occluded facial parts in an inherent and automatic manner.

Existing studies on utilizing deep learning for FEA under occlusion are still very limited, and they mainly focus on using a single deep learning architecture. Toser *et al.* [2016] employed a 8-layer CNN for AU detection on 3D facial data with self-occlusion caused by large pose variations. The CNN accepted as input similarity normalized images and combined gradient descent with selective methods to improve the convergence of its optimization. An augmented data was created based on the BP4D database using 3D face information and renderings of the face with different rotations, yielding faces with head poses ranging (-18, 90) degrees for yaw and (-54, 54) degrees for pitch. The CNN produced a mean F1 score of 0.55 for recognizing 11 AUs on the augmented data. Cheng *et al.* [2014] presented a deep structure for FER under partial occlusion. Multi-scale and multi-orientation Gabor magnitudes were extracted from facial images using Gabor filters, and were taken as input to a three-layer Deep Boltzmann Machine (DBM) for emotion classification. The DBM had a structure of 3,500-600-600-7 elements and incorporated a pre-training and a fine-turning process to learn the best weights for encoding the structure of the face including occlusion. The impact of occlusion was then reduced by compressing the features from occluded parts in the DBM. The DBM was evaluated for classifying six basic and neutral emotions from JAFFE images with occlusion simulated by artificially added black bars. It achieved 85.7% and 77.1% for non-occlusion and occluded upper part of the face, as well as the same accuracy of 82.9% for occluded eyes, mouth and lower part of the face. Ranzato *et al.* [2011] presented a deep generative model that uses a gated Markov Random Field (MRF) as the first layer of a DBN for FER under occlusion. The MRF learnt the covariance structure and intensities of image pixels, while the DBN included several layers of Bernouilli hidden variables to model the statistical structure in the hidden activities of the MRF. The missing pixels in facial occlusion were filled in by propagating the occluded image through all layers in the model using a sequence of posterior distributions learnt from training data. From the reconstructed facial region, features were extracted and used for FER with a linear classifier. Occlusions of the eyes, mouth, nose, right/bottom/top halves, and 70% of the pixels at random were artificially generated based on the CK and Toronto Face (TF) databases. The results showed a higher overall accuracy than a Gabor-based, sparse coding, SVM-based approaches. However, the deep generative model assumes prior knowledge of the location of occlusion, requires large training data, and is computationally expensive.

Deep learning techniques also showed state-of-the-art performance in the FERA 2017 challenge. The challenge used a dataset from the BP4D database having six intensity levels of seven AUs and nine head poses, as described in Section 3. Batista *et al.* [2017] presented a unified CNN (AUMPNet) to detect AUs and estimate their intensity simultaneously from multi-view facial images through a multi-task loss. The pose estimate was added to the multitask loss to generate features invariant to head pose changes. The AUMPNet demonstrated better performance than the FERA 2017 baseline, and it achieved mean F1 scores of 0.521 and 0.506 for AU detection, and mean ICCs of 0.499 and 0.399 for intensity estimation on the validation and test subsets, respectively. Zhou *et al.* [2017] proposed a Multi-Task Deep Network (MTDN) for AU intensity estimation. Three pose-dependent AU regressors and one pose estimator were trained, and they shared the same bottom layers of a deep CNN. The final AU estimate was taken as the dot product between a winning pose regressor and the output of the pose estimator. The MTDN network achieved a mean RMSE of 0.823, ICC of 0.601, and Pearson Correlation Coefficient (PCC) of 0.620 on estimating AU intensity on the validation subset, as well as RMSE of 0.879 and ICC of 0.446 on the test subset. Tang *et al.* [2017] fine-tuned the VGG network for AU detection from multi-view facial regions that were segmented from facial images using morphology operations. The network achieved F1 of 0.574 and accuracy of 0.778 on the test subset. Li *et al.* [2017a] proposed a multi-AUs late fusion approach. The hand-crafted LBP-TOP features and automatically extracted CNN features were used separately for training three AU classifiers. The predictions of all classifiers were concatenated and taken as the input of a second-level AU classifier. The approach achieved F1 of 0.498 and accuracy of 0.694 on the test subset. Similarly, state-of-the-art results have also been observed for deep learning techniques in the EmotiW 2015 and 2016, where improved versions of RNN, CNN, and LSTM were reported as the top performers in video based, image based, and group-level emotion recognition.

The power of deep learning techniques is that they can automatically learn the most discriminative feature patterns of facial expressions from the raw face data. They normally do not require a separate process of occlusion detection or reconstruction. Specifically, the occlusion information can be inherently embedded in the feature set automatically learnt by deep learning architectures. Given the state-of-the-art performance of deep learning techniques in various compute vision tasks, we can anticipate that deep learning will potentially be one of the most effective approaches to handling occlusion for FEA. However, the use of deep architectures for FEA has to overcome several constraints such as the need of a large amount of training data to ensure proper feature learning, the difficulty of tuning a large number of system parameters, and the requirement of expensive computation.

### 4.7 Summary of FEA Approaches

The state of the art on automatic FEA approaches under partial occlusion can be summarized as follows.

1) The primary types of occlusion include occluded eyes, mouth, left/right, upper/lower face, random placed blocks, hands, glasses, noise (as shown in Fig. 11), and self-occlusion caused by head pose variations. Few studies have tested occlusion in brows [Miyakoshi and Kato 2011] or arising from hair [Xia, et al. 2009] and blur [Tan Dat and Ranganath 2008],[Tan Dat and Surendra 2008], as well as missed frames [Yongmian and Qiang 2005].

2) The majority of existing studies focus on artificial occlusion simulated by removing occlusion related features or by superimposing graphically generated masks into a certain region of the face. Few attempts have been made toward using naturally occurring occlusion from real-life data, such as mouth masks and sunglasses [Shuai-Shi, et al. 2013], and hand occlusion [Yongmian and Qiang 2005],[Tan Dat and Ranganath 2008],[Tan Dat and Surendra 2008].

3) Most existing evaluations are based on the JAFFE, CK, and CK+ databases with non-occluded faces. Some initial efforts were reported on real-life data with natural occlusion arising from sunglasses, medical mask [Shuai-Shi, et al. 2013] and hands [Yongmian and Qiang 2005], but the results were primarily used for demonstrating the performance of the system on sample occluded data and no thorough evaluation outcomes such as emotion classification accuracy were reported on the whole dataset. No work has been found to investigate FEA under occlusion from 3D face data.

4) All existing studies focus on a single type of occlusion in frontal faces, and to our best knowledge, the presence of co-existed occlusion or multiple faces has not been investigated yet.

5) Most work extracts features directly from occluded facial images without incorporating a pre-processing step of occlusion detection. Only few attempts [Xia, et al. 2009],[Huang, et al. 2012],[Lin, et al. 2013] have investigated occlusion detection techniques and integrated them in complete FEA systems. Most approaches require accurate face location and alignment, and robust facial feature tracking.

6) Features used are largely restricted to texture or geometry from the 2D visual face modality only, and few studies have investigated fusion of them [Huang, et al. 2012] and utilized skin color features [Lin, et al. 2013]. Existing works on 3D data primarily focus on self-occlusion generated by varying head poses of a 3D face model. Only few studies have been found on exploiting the fusion of features from multiple modalities.

7) Most studies place emphasis on the six basic emotions plus neutral. Only few studies have been reported on exploiting non-basic emotions, such as contempt [Huang, et al. 2012], grammatical expressions [Tan Dat and Ranganath 2008],[Tan Dat and Surendra 2008] and AUs [Kapoor, et al. 2003],[Lin, et al. 2013],[Yongmian and Qiang 2005]. No occlusion-specific work has been found on using dimensional spaces.

8) Although various types of feature descriptors have been used in existing work, there is yet an agreement on the most effective feature descriptor for handling facial occlusion. For instance, which descriptor works best with the sparse representation is still largely an unanswered question? From this perspective, it seems that deep learning presents a unique advantage by automatically leaning the most discriminative features for FEA.

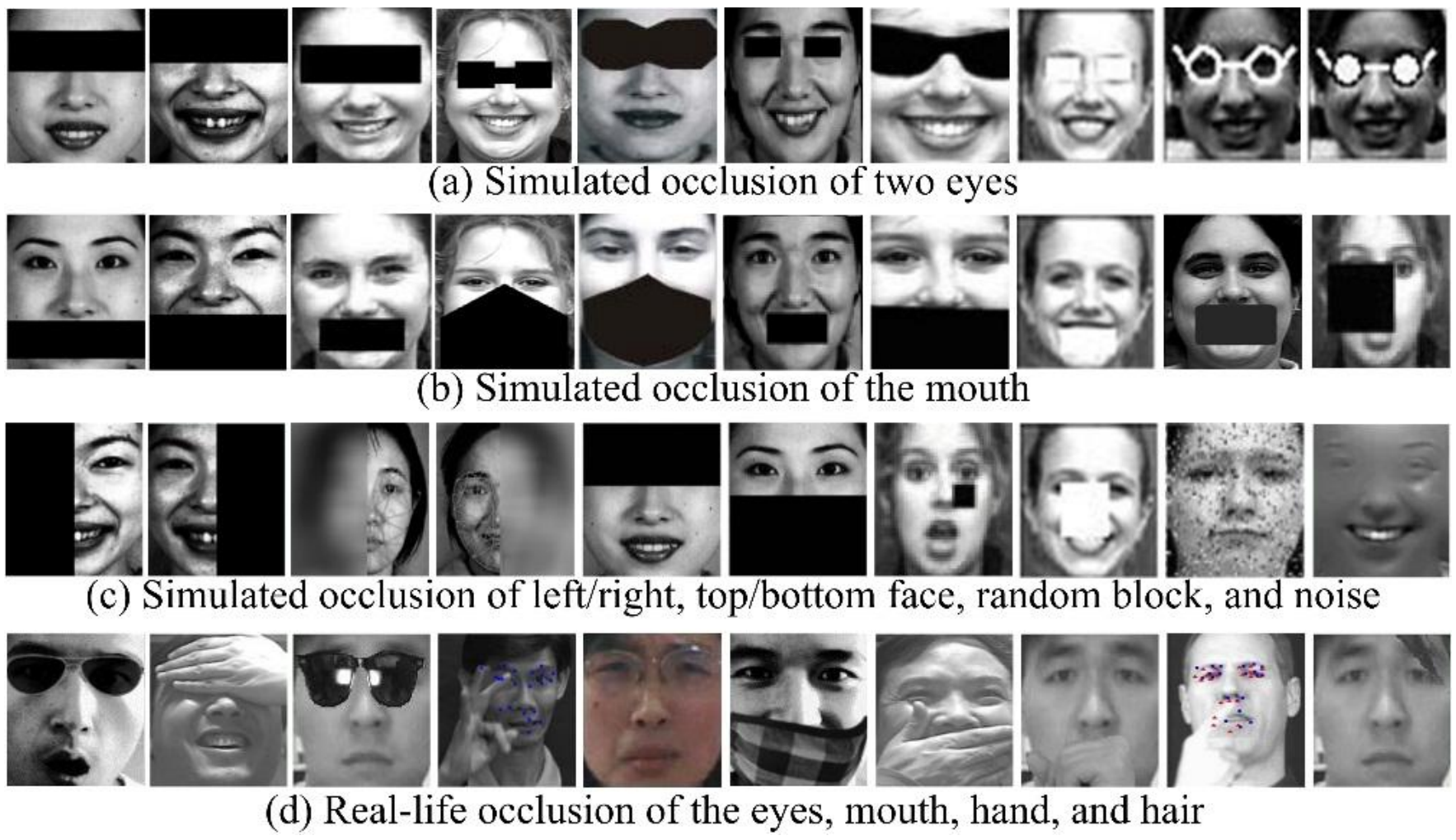

**Fig. 11. Illustration of various types of facial occlusion in existing FEA studies.**

## 4.8 Top Five Benchmark Approaches

To facilitate direct performance comparisons of future algorithms with the state-of-the-art results, we identified the top five approaches from existing studies on the JAFFE and CK/CK+ databases respectively. The criterion for selecting these approaches is that they achieved the so-far highest overall accuracies of classifying six basic and neutral emotions on the two databases between 2001 and 2016. In addition, we also included recent approaches from the FERA 2017 challenge. Table 4 lists the accuracies of these approaches under different types of facial occlusion. It should be noted that the accuracies on the JAFFE and CK/CK+ databases provide a simple indicator of the performance and may not be directly comparable due to the differences in train-test strategies, face pre-processing steps, occlusion simulation methods, etc. in these approaches.

Table 4. Accuracy (%) of Top Five Benchmark Approaches under Different Types of Facial Occlusion.

(a) the JAFFE Database

| Ref. | Approach | Face size | Emo No. | Occlusion | | | | | | | |
|---|---|---|---|---|---|---|---|---|---|---|---|
| | | | | Non | Eyes | Mouth | Brow | Upper | Lower | Block | Glass |
| [Cotter 2011] | FLSRC | 96×72 | 7 | 96.7 | - | 93.4 | - | - | - | 85.0 | - |
| [Cotter 2010b] | WVSRC | 96×72 | 7 | 95.3 | - | - | - | 63.8 | 77.0 | 86.4 | - |
| [Cheng, et al. 2014] | DBM | 80×70 | 7 | 85.7 | 82.9 | 82.9 | - | 77.1 | 82.9 | - | - |
| [Kotsia, et al. 2008] | DNMF | 80×60 | 6 | 85.2 | 82.5 | 81.5 | - | - | - | - | - |
| [Zhang, et al. 2014b] | Gabor Temp. | 48×48 | 7 | 81.2 | 80.3 | 78.4 | - | - | - | 48.8 | 75.1,79.8 |

(b) the CK/CK+ Database

| Ref. | Approach | Face size | Emo. No. | Occlusion | | | | | | | |
|---|---|---|---|---|---|---|---|---|---|---|---|
| | | | | Non | Eyes | Mouth | Nose | Lower | Block | Noise | Glass |
| [Zhang, et al. 2014b] | Gabor Temp. | 48×48 | 7 | 95.3 | 95.1 | 90.8 | - | - | 75 | - | 91.5,95.0 |
| [Huang, et al. 2012] | CFD-OD-WL | orig | 7 | 93.2 | 93.0 | 79.1 | - | 73.5 | - | 79.7 | - |
| [Zhi, et al. 2011] | GSNMF+NN | 60×60 | 6 | - | 93.3 | 91.4 | 94.0 | - | - | - | - |
| [Kotsia, et al. 2008] | Shape+SVM | 80×60 | 6 | 91.4 | 88.4 | 86.7 | - | - | - | - | - |
| [Ouyang, et al. 2013] | LBPM+SRC | 64×64 | 6 | - | 72.4 | - | - | - | - | 87.0 | - |

(c) the BP4D Database (FERA 2017 Challenge)

| Ref. | Approach | Test size | AU No. | Inten. level | Pose no. | AU recognition | | | Intensity estimation | | |
|---|---|---|---|---|---|---|---|---|---|---|---|
| | | | | | | F1 | Acc. | 2AFC | RMSE | ICC | PCC |
| [Zhou, et al. 2017] | MTDN | 1080 V | 7 | 6 | 9 | - | - | - | 0.879 | 0.446 | - |
| [Batista, et al. 2017] | AUMPNet | 1080 V | 7 | 6 | 9 | 0.506 | - | - | - | 0.399 | - |
| [Li, et al. 2017a] | multi-AU fusion | 1080 V | 7 | 6 | 9 | 0.498 | 0.694 | - | - | - | - |
| [Tang, et al. 2017] | VGG | 1080 V | 7 | 6 | 9 | 0.574 | 0.778 | - | - | - | - |
| [Amirian, et al. 2017] | sparse coding | 1080 V | 7 | 6 | 9 | - | - | - | 0.970 | 0.295 | - |
| [Valstar, et al. 2017] | CRF or CORF | 1080 V | 7 | 6 | 9 | 0.452 | 0.561 | 0.537 | 1.403 | 0.217 | 0.221 |

Note: '-' means values not available, the facial region in [Miyakoshi and Kato 2011] is set to 30×30 pixels between two eyes, and Ref. [Valstar, et al. 2017] is the baseline approach for the FERA 2017 Challenge. Abbreviations: CRF – Conditional Random Field (AU occurrence sub-challenge), CORF – Conditional Ordinal Random Field (intensity sub-challenge), DBM - Deep Boltzmann Machine, VGG - Oxford Visual Geometry Group network.

## 5 Effect of Occlusion on Facial Expression

Investigations into the effect of occlusion on the classification performance of facial expressions can provide useful insights into the most informative facial parts for an expression, and thus can be useful for designing FEA systems. Existing investigations are from either computer vision or human perception.

### 5.1 Computer Vision Investigation

Computer vision investigations are generally based on the recognition performance of FEA systems. Buciu *et al.* [2005] showed that occlusion of the eyes and mouth have a similar effect on the overall performance on the JAFFE database, while occlusion of the mouth exerts a larger effect than occluded eyes on the CK database. Occluded eyes and mouth affect the most sadness and neutral respectively on the JAFFE database, and both anger and sadness on the CK database. Buciu *et al.* [Kotsia, et al. 2008] also conducted a machine experiment and a human observer experiment (2 experts and 13 non-experts) on the JAFFE and CK databases. The results demonstrated that an occluded mouth results in a decrease of more than 50% in the overall classification accuracy than occluded eyes, indicating a more important role of the mouth than the eyes for FER. Occlusion on the left or right side of the face has little effect on the accuracy. An occluded mouth affects the classification accuracies of anger, fear, happiness and sadness more, while an occluded eye affects those of disgust and surprise more. The results are in line with those from human observers, indicating the consistency between computer vision and human vision in discriminating facial expressions from occluded faces. Azmi and Yegane [Azmi and Yegane 2012] investigated the effect of occlusion of the mouth, eyes, upper and lower parts of the face. Occluded eyes were found to affect the overall accuracy more than an occluded mouth. Occluded eyes have more effect on the classification accuracies of surprise, anger and sadness, while an occluded mouth affects more on those of disgust and fear. Upper face occlusion affects the classification of sadness, surprise and anger most, while lower face occlusion impacts more that of disgust, fear and neutral. Regarding train-test strategies, studies [Ranzato, et al. 2011], [Zhang, et al. 2014b] found that using both training and test data with occlusion produces higher accuracy than using non-occluded training data and occluded testing data. This indicate the importance of informing the learning model of the presence of occlusion patterns at the training stage. From the above results, we can find that there is no direct agreement on a more important role between the mouth and eyes. Take the JAFFE database as an example, some studies [Buciu, et al. 2005] found an equally important role between them. By contrast, some studies [Kotsia, et al. 2008] found that the mouth is more important, while some [Shuai-Shi, et al. 2013] indicated that the eyes are more important. One primary reason for the contrary results is due to different computer vision systems used in those work. Those systems may vary dramatically in aspects such as data pre-processing step, feature type, classification algorithm, system setting, and train-test strategy. These differences can have direct big impact on the results of the systems even when the same database with the same type of occlusion is used for evaluation.

Table 5 presents a summary of the reductions in accuracy owing to facial occlusion in comparison with non-occlusion for six basic emotions plus neutral, and the overall performance. Almost all present investigations are based on the JAFFE and CK databases. For studies that have not provided the reductions in accuracy, they are not listed.

From Table 5, we can see that:

- Occlusion of the mouth results in more reduction in accuracy than occlusion of eyes for anger, fear, happiness, sadness, and the overall performance.
- For disgust, occlusion of eyes leads to more reduction in accuracy than occlusion of the mouth on the CK database, but less reduction on the JAFFE database.
- Occlusion of the lower face causes significant reduction of the classification accuracies of six basic emotions plus neutral, and of the overall accuracy.
- Occlusion of the upper face produces large reduction in accuracy for anger and happiness, but small reduction for neutral and fear.
- Occlusion of the nose leads to little reduction in accuracy for all six basic emotions.
- Occlusion of random blocks causes large reduction of the overall accuracy.

## 5.2 Human Perception Investigation

Human perception investigations often asked recruited subjects to identify pre-defined facial expressions from a face with certain parts occluded. The results reveal the visual information that is perceptually necessary and sufficient in the human recognition. Instead of providing a comprehensive review on previous work (which is not the focus of this survey), we introduce several typical work.

Early psychological studies [Dunlap 1927],[Ruckmick 1921] focused on the question of whether there is one facial area which can best distinguish among facial expressions. This question was later largely answered by the predominant evidence from researchers such as Ekman [Boucher and Ekman 1975],[Ekman *et al.* 2013] and Hanawalt [Hanawalt 1942] that the most distinctive facial component varies with each emotion. Using static photographs of posed facial expressions, it was generally found that the most important facial components are mouth/cheeks, eyes/eyelids, and brows/forehead, and that disgust is best distinguished from the mouth, fear from the eyes, sadness from both brows and eyes, happiness from both mouth and eyes, anger from mouth and brows, and surprise from all and eyes, anger from mouth and brows, and surprise from all three components.

Recent studies on investigating the effect of occlusion on human perception tend to use spontaneous facial expressions, video sequences, and subjects with different ages. Halliday [Halliday 2008] asked 56 female participants to identity happiness, sadness and fear from static photographs of genuine and posed facial expressions from a single displayer, with four regions occluded: the forehead and eyebrows, the nose and cheeks, the eyes, and the mouth. The results revealed that participants can accurately identify emotions from limited information, and the mouth and eyes are the two most crucial regions for recognizing genuine emotions. Bassili [Bassili 1979] found that the participants more accurately recognize six basic emotions using temporal displays in video sequences than static displays of peak emotions, implying the importance of facial motion in assisting the recognition. The bottom part of the face is able to generate a higher overall recognition rate than the top part. Using 11 participants, Nusseck [Nusseck *et al.* 2008] showed that the fusion of the mouth, eyes and eyebrows regions is sufficient to generate acceptable results for most of nine conversational expressions. By comparing the recognition rates of facial expressions with simulated sunglasses or masks between younger children, children and adult students, Roberson *et al.* [Roberson *et al.* 2012] observed that the capacity of accurate decoding of facial expressions under eyes and mouth occlusion grows slowly with ages, but the inversion of sunglasses does not affect the performance of 5–6 year olds.

Table 5. Summary of Accuracy Deduction (% in Parentheses) for Each Emotion and Overall Performance due to Facial Occlusion.

| | Database | Approach | Occlusion | | | | | | |
|---|---|---|---|---|---|---|---|---|---|
| | | | **Non** | **Eyes** | **Mouth** | **Upper** | **Lower** | **Nose** | **Random** |
| AN | CK | Gabor [Kotsia, et al. 2008] | 82 | 79.3(-2.7) | 70(**-12**) | - | - | - | - |
| | | DNMF [Kotsia, et al. 2008] | 78.3 | 74(-4.3) | 69.3(-9) | - | - | - | - |
| | | SVM [Kotsia, et al. 2008] | 86.9 | 82(-4.9) | 80.7(-6.2) | - | - | - | - |
| | | GSNMF [Zhi, et al. 2011] | 98.7 | 94.7(-4) | 94(-4.7) | - | - | 97.3(-1.4) | - |
| | | PCA [Towner and Slater 2007] | 94 | - | - | 70(**-24**) | 48(**-46**) | - | - |
| | JAFFE | LGBP [Azmi and Yegane 2012] | 99.7 | 85(**-14.7**) | 99.7(-0) | 89(-9.7) | 96(-3.7) | - | - |
| | | Bayesian [Miyakoshi and Kato 2011] | 63.3 | 70(+6.7) | 46.7(**-16.6**) | - | - | - | - |
| DI | CK | Gabor [Kotsia, et al. 2008] | 94.1 | 81.5(**-12.6**) | 85.1(-9) | - | - | - | - |
| | | DNMF [Kotsia, et al. 2008] | 82 | 77.8(-4.2) | 80(-2) | - | - | - | - |
| | | SVM [Kotsia, et al. 2008] | 86.7 | 83.8(-2.9) | 85.2(-1.5) | - | - | - | - |
| | | GSNMF [Zhi, et al. 2011] | 93.3 | 91.3(-2) | 91.3(-2) | - | - | 92.7(-0.6) | - |
| | | PCA [Towner and Slater 2007] | 100 | - | - | 94(-6) | 60(**-40**) | - | - |
| | JAFFE | LGBP [Azmi and Yegane 2012] | 94.8 | 91.3(-3.5) | 87.5(-7.3) | 94.2(-0.6) | 83.3(**-11.5**) | - | - |
| | | Bayesian [Miyakoshi and Kato 2011] | 55.2 | 48.3(-6.9) | 31(**-24.2**) | - | - | - | - |
| FE | CK | Gabor [Kotsia, et al. 2008] | 93 | 92.5(-0.5) | 87.2(-5.8) | - | - | - | - |
| | | DNMF [Kotsia, et al. 2008] | 76 | 74(-2) | 71(-5) | - | - | - | - |
| | | SVM [Kotsia, et al. 2008] | 92.9 | 91.9(-1) | 87.3(-5.6) | - | - | - | - |
| | | GSNMF [Zhi, et al. 2011] | 95.3 | 94(-1.3) | 88(-7.3) | - | - | 94(-1.3) | - |
| | | PCA [Towner and Slater 2007] | 82 | - | - | 78(-4) | 70(**-12**) | - | - |
| | JAFFE | LGBP [Azmi and Yegane 2012] | 93 | 86.8(-6.2) | 80.7(**-12.3**) | 92.7(-0.3) | 86.8(-6.2) | - | - |
| | | Bayesian [Miyakoshi and Kato 2011] | 71.9 | 56.3(**-15.6**) | 53.1(**-18.8**) | - | - | - | - |
| HA | CK | Gabor [Kotsia, et al. 2008] | 90.6 | 88.2(-2.4) | 83.2(-7.4) | - | - | - | - |
| | | DNMF [Kotsia, et al. 2008] | 96.5 | 95(-1.5) | 93.1(-3.4) | - | - | - | - |
| | | SVM [Kotsia, et al. 2008] | 95.7 | 93.6(-2.1) | 90.9(-4.8) | - | - | - | - |
| | | GSNMF [Zhi, et al. 2011] | 96.7 | 96(-0.7) | 93.3(-3.4) | - | - | 94(-2.7) | - |
| | | PCA [Towner and Slater 2007] | 86 | - | - | 73(**-13**) | 67(**-19**) | - | - |
| | JAFFE | LGBP [Azmi and Yegane 2012] | 95.6 | 89.1(-6.5) | 97.8(+2.2) | 89.1(-6.5) | 95(-0.6) | - | - |
| | | Bayesian [Miyakoshi and Kato 2011] | 78.1 | 78.1(-0) | 56.3(**-21.8**) | - | - | - | - |
| SA | CK | Gabor [Kotsia, et al. 2008] | 93 | 88.5(-4.5) | 87.4(-5.6) | - | - | - | - |
| | | DNMF [Kotsia, et al. 2008] | 90.4 | 89.3(-1.1) | 88.7(-1.7) | - | - | - | - |
| | | SVM [Kotsia, et al. 2008] | 89.5 | 86.7(-2.8) | 82.8(-6.7) | - | - | - | - |
| | | GSNMF [Zhi, et al. 2011] | 95.3 | 94.7(-0.6) | 93.3(-2) | - | - | 94.7(-0.6) | - |
| | | PCA [Towner and Slater 2007] | 84 | - | - | 82(-2) | 74(**-10**) | - | - |
| | JAFFE | LGBP [Azmi and Yegane 2012] | 95.8 | 95.6(-0.2) | 95.8(-0) | 89.1(-6.7) | 95.8(-0) | - | - |
| | | Bayesian [Miyakoshi and Kato 2011] | 63.3 | 66.7(+3.4) | 40(**-23.3**) | - | - | - | - |
| SU | CK | Gabor [Kotsia, et al. 2008] | 96.7 | 90.8(-5.9) | 93.3(-3.4) | - | - | - | - |
| | | DNMF [Kotsia, et al. 2008] | 97 | 94.8(-2.2) | 95.1(-1.9) | - | - | - | - |
| | | SVM [Kotsia, et al. 2008] | 96.8 | 92.4(-4.4) | 93.3(-3.5) | - | - | - | - |
| | | GSNMF [Zhi, et al. 2011] | 92.7 | 89.3(-3.4) | 88.7(-4) | - | - | 91.3(-1.4) | - |
| | | PCA [Towner and Slater 2007] | 99 | - | - | 99(-0) | 83(**-16**) | - | - |
| | JAFFE | LGBP [Azmi and Yegane 2012] | 96.3 | 75.3(-21) | 90(-6.3) | 78.7(**-17.6**) | 93.3(-3) | - | - |

| | | | | | | | | |
|---|---|---|---|---|---|---|---|---|
| | | Bayesian [Miyakoshi and Kato 2011] | 90 | 83.3(-6.7) | 70(**-20**) | - | - | - | - |
| NE | JAFFE | LGBP [Azmi and Yegane 2012] | 99 | 98.7(-0.3) | 99(-0) | 98.7(-0.3) | 74.3(**-24.7**) | - | - |
| Overall | CK | Gabor [Kotsia, et al. 2008] | 91.6 | 86.8(-4.8) | 84.4(-7.2) | - | - | - | - |
| | | DNMF [Kotsia, et al. 2008] | 86.7 | 84.2(-2.5) | 82.9(-3.8) | - | - | - | - |
| | | SVM [Kotsia, et al. 2008] | 91.4 | 88.4(-3) | 86.7(-4.7) | - | - | - | - |
| | | CFDWL [Huang, et al. 2012] | 93.2 | 93(-0.2) | 79.1(**-14.1**) | - | 73.5(**-19.7**) | - | 86.8(-6.4) |
| | | Gabor [Zhang, et al. 2014b] | 95.3 | 95.1(-0.2) | 90.8(-4.5) | - | - | - | 75(**-20.3**) |
| | | MCC [Buciu, et al. 2005] | 93.6 | 87.2(-6.4) | 92.3(-1.3) | - | - | - | - |
| | | PCA [Towner and Slater 2007] | 75, 85 | - | - | 70(-5) | 82(-3) | - | - |
| | | SRC [Ouyang, et al. 2013] | 97.7 | 72.4(**-25.3**) | - | - | - | - | - |
| | JAFFE | Gabor [Kotsia, et al. 2008] | 88.1 | 83.1(-5) | 81.5(-6.6) | - | - | - | - |
| | | DNMF [Kotsia, et al. 2008] | 85.2 | 82.5(-2.7) | 81.5(-3.7) | - | - | - | - |
| | | LGBP [Azmi and Yegane 2012] | 96.3 | 88.8(-7.5) | 92.8(-3.5) | 90.2(-6.1) | 89.2(-7.1) | - | - |
| | | LGBPHS [Shuai-Shi, et al. 2013] | 93.4 | 85.5(-7.9) | 92.1(-1.3) | - | - | - | - |
| | | Gabor [Zhang, et al. 2014b] | 81.2 | 80.3(-0.9) | 78.4(-2.8) | - | - | - | 48.8(**-32.4**) |
| | | MCC [Buciu, et al. 2005] | 89.7 | 83.5(-6.2) | 84(-5.7) | - | - | - | - |
| | | Bayesian [Miyakoshi and Kato 2011] | 70.3 | 67.1(-3.2) | 49.5(**-20.8**) | - | - | - | - |
| | | RPCA [Xia, et al. 2009] | 87.5 | 68.8(**-18.7**) | - | - | - | - | - |
| | | DBN [Cheng, et al. 2014] | 85.7 | 82.9(-2.8) | 82.9(-2.8) | 77.1(-8.6) | 82.9(-2.8) | - | - |

Note: '-' means values not available, and all accuracy reductions (with reference to non-occlusion) larger than 10% are highlighted in bold. Some approaches have higher accuracies under occlusion than non-occlusion. Random block occlusion accounts for a quarter of the face area, and Ref. [Xia, et al. 2009] used a fusion set of JAFFE and BHUFE images.

Abbreviations: CFD - Combination of component-based Facial expression representation and fusion module, CFDWL - CFD based on Weight Learning, DBN – Deep Belief Network, DNMF- Discriminant Non-negative Matrix Factorization, GSNMF - Graph-preserving Sparse Non-negative Matrix Factorization, LGBP - Local Gabor Binary Pattern, LGBPHS - LGBP Histogram Sequence, MCC- Maximum Correlation Classifier.

## 5.3 Summary of Effect of Occlusion

From the above analysis, we can observe that most current investigations focus on 1) six basic emotions and neutral, 2) occlusion of the mouth, eyes, left or right, upper or lower parts of the face, and random blocks, 3) JAFFE and CK databases, and 4) emotional stimuli from actors/actresses. Few studies have investigated non-basic emotions such as contempt [Huang, et al. 2012], and occlusion of noise [Bourel, et al. 2002], nose [Zhi, et al. 2011] and real-life sunglasses [Shuai-Shi, et al. 2013].

Table 6 summarizes the effect of six types of occlusion on the six basic emotions plus neutral, as well as the overall performance based on some of existing investigations on computer vision and human perception. The results are listed to provide readers a first impression on the effect, as an exhaustive survey on this field is out of the scope of this paper. It can be seen that:

- The left and right halves of the face equally effect the overall performance [Bourel 2001],[Kotsia, et al. 2008],[Bourel, et al. 2002],[Shuai-Shi, et al. 2013].
- The mouth is the most important facial region and an occluded mouth has large effect on the classification of six basic emotions and the overall performance, but relatively small effect on that of neutral.
- The eyes are the second most important facial region, and occluded eyes have large effect on the classification of sadness, disgust, and surprise, but only small effect on that of happiness and neutral.
- The upper face has more impact on the classification of anger [Bourel, et al. 2002],[Towner and Slater 2007],[Azmi and Yegane 2012], but small impact on the overall performance [Bourel 2001] and classification of fear.
- Occlusion of the lower face significantly impacts the overall performance [Towner and Slater 2007],[Huang, et al. 2012].
- Occlusion of the nose has large effect on the classification of sadness, but small effect on the overall performance.
- Occlusion of the eyebrows have small effect on classification of happiness, sadness, fear and disgust.

Overall, there is no absolute consensus on the effect of facial occlusion in both computer vision and human perception experiments. Occlusion from the same facial part may exert different effects on the classification of the same emotion, which implies the effect is largely context-dependent, and impacted by local factors specific to a particular experiment, such as features, classifiers, participants, emotional stimuli and evaluation protocols.

Table 6. Summary of Effect of Occlusion on Classification of Six Basic Emotions + Neutral and Overall Performance.

| | Eyes | Mouth | Upper | Lower | Nose | Brow |
|---|---|---|---|---|---|---|
| AN | --++++ | ---++++ | ++ | + | + | - |
| DI | --+++++ | -+++++ | N/A | -++ | + | -- |
| FE | --++ | -++++++ | - | --+ | -+ | -- |
| HA | ---+ | ---++++++ | N/A | + | + | --- |
| SA | -+++++ | -++++++++ | -+ | - | ++ | --- |
| SU | --+++++ | --++++ | -+ | - | N/A | -+ |
| NE | - | ++ | N/A | + | N/A | N/A |
| Overall | +++ | ++++ | - | ++++ | -- | N/A |

Note: '+' and '-' indicate large and small effect respectively. Multiple '+' or '-' represents the number of studies with a large or small effect. Abbreviation: N/A – Not Available.

# 6 Challenges and Opportunities

Automatic FEA in partially occluded faces is a field that is just at its very beginning stage and has received relatively less investigations previously. Thus, it is extremely important to discuss existing challenges that form the major obstacles to the current progress and possible opportunities that should be paid special attention in promoting the future work. This section presents key unresolved issues identified in Sections 4 and 5, and discuss possible solutions or opportunities to address these issues. We limit our focus to only those issues closely related to handling facial occlusion, and for uncovered issues with respect to face location, face normalization, face tracking, feature extraction and classifier design in non-occluded faces, readers are referred to [Pantic and Rothkrantz 2000],[Fasel and Luettin 2003],[Gunes, et al. 2011],[Gunes and Schuller 2013].

## 6.1 Database Creation and Labeling

Most current studies are based on JAFFE, CK or CK+ databases without occluded faces, and are restricted to a limited number of single types of artificially generated occlusion and to facial expressions of six basic emotions plus neutral. This is largely due to the lack of comprehensive benchmark datasets that include a dense set of various types of frequent natural facial occlusion and well annotated labels of facial expressions. The creation of FEA databases with facial occlusion is a more complicated and time-consuming work than that without occlusion, and it has to overcome the following issues:

1) *Decision on what kinds of facial occlusion.* It is generally agreed that occlusion from the most informative facial regions, such as the mouth, eyes, eyebrows and nose, should be included. For a specific application, a certain part of the face may become crucially important and needs to be considered individually and included as well. Even the facial parts are determined, it is still unclear what types of specific objects should be utilized to occlude these parts and to what extent the parts should be occluded in the procedure of occlusion simulation. For a given facial part, there may exist a wide range of specific objects for selection and these objects may have substantially varied properties such as color, shape, size, appearance, component and material. For instance, occlusion of the eyes can be simulated by asking subjects taking different types of eyeshades or glasses, such as sunglasses, vision correction glasses, goggles, protective glasses, and specified glasses (e.g. Google glasses). Whether a type of occlusion should be simulated by objects with the same property or different properties, and whether the co-occurrence of multiple occlusion should be considered remain questions. Although it is generally believed that objects with varied visual and physical properties should be included to simulate real-life scenarios as closely as possible, the varieties presented in the occlusion may pose a big challenge for effective training and tests of the FEA algorithms.

2) *Collection of spontaneous expressions of facial emotions under occlusion.* Human facial expression is a complex process, involving psychological activities, cognitive understanding and physical behaviors that come together to create the subjective experience. It is relatively easy to elicit some prototypical emotions such as happiness, sadness and surprise, from subjects by natural face-to-face communications or showing them proper emotional stimuli. However, it becomes increasingly difficult when moving beyond prototypical emotions to other uncommon context-dependent expressions that are seldom used in the normal life and involve small subtle changes in facial components, such as contempt, curiosity and attentiveness, particularly in the presence of facial occlusion. The occlusion that is either artificially superimposed on the face or naturally occurring may significantly interfere subjects' spontaneous reactions to emotional stimuli and influence their natural ways of expressing facial expressions. This is because subjects may need a certain amount of time to get used to the presence of occlusion, to accurately express facial movements, and to fully engage themselves to the emotional stimuli and the contextual environment. Thus, the interference from occlusion may severely influence the reliability and accuracy of elicited facial expressions.

Except for the common issues such as participant recruitment, participation agreement, ethics and copyright, there are extra issues and protocols that should be considered during spontaneous facial expression elicitation and data collection. First, the metadata of the database, such as the types of both occlusion and emotions, the number of subjects and the data modality, should be determined based on the specific aims, tasks, or applications of the data collection. However, as a starting point for current academic research, it may be a good idea to include the commonly used types such as six basic emotions, and occlusions of the mouth, eyes, nose, top half, and bottom half of the face. While it is often beneficial to include as many subjects as possible, the number of subjects participated should be controlled to a reasonable level which fits well to the resources available to the data collection project such as budget, time, equipment, staff, etc. To avoid the necessity of artificially adding occlusion to the face and to reduce the impact of occlusion on the elicited facial expressions, it is advisable to select only those subjects who have a specific type of occlusion in their normal lives. However, this will greatly limit the number of subjects and the types of occlusion that can be collected in real practice. With respect to data modality, decisions should be made on whether the data should cover audio vs. visual modality, visual vs. thermal modality, 2D vs. 3D data, static images vs. video sequences, body gestures, single vs. multiple faces, single vs. multiple views, etc. or combinations of them. Next, to elicit facial expressions from subjects as spontaneously as possible, it is essential to select proper simulating materials that can be relatively easier to arouse natural affective responses from the subjects. The simulating methods can be story-telling, watching videos, playing games, attending social events, or face-to-face communications, etc. Another important aspect is to provide a natural and relax environment, where the participants are allowed to freely express their feelings and emotions ideally without any constraints on their activities, movements, positions, gestures, and taking on or off the occlusion. Dependent on the specific requirement of the project, the participants may or may not be told details about the collection procedure such as where, when, how and how long their affective responses will be collected. For the purpose of comparing differences in a subject's affective responses, it may be also worth collecting two separate sets of data for the same subject - one before and the other after the subject is informed of those details. After data collection, the collected data undergoes normal procedures such as post-processing, emotion annotation, experimental validation, and finally may be made accessible by the public.

Spontaneous facial expression collection is not an easy task. To improve the accuracy of the elicited expressions and label annotation, it is advisable to take into account a list of creative strategies recommended by psychological studies for inducing emotions, particularly those subtle and contextual-dependent ones [Zeng *et al.* 2009]. However, this may increase the difficulty of subject recruitment and greatly limit the type of occlusion that can be collected. The data collection process is also largely hindered by the lack of awareness by engineers who are actually in charge of the whole procedure. Rather than collecting spontaneous facial expressions in a controlled laboratory environment, recent studies [Benitez-Quiroz *et al.* 2016], [Benitez-Quiroz *et al.* 2017] have shifted to collecting a large number of images of facial expressions with associated emotion keywords from the Internet. This approach may also be used for collecting occluded facial images with spontaneous expressions in 'wild' environments.

3) *No criterion for how the occlusion should be annotated.* Once the data of facial expressions with occlusion was recorded, which properties of occlusion should be annotated and how to annotate them become a real challenge because the occlusion may be the results of a wide range of objects with different properties. It is generally accepted that the location of occlusion is the most important property that has the biggest impact on FEA, which can be annotated by pixelwise labels of occluded regions. However, whether other properties of the occlusion, such as the specific type (e.g. sunglasses or eyeglasses), the intensity of occlusion (e.g. 50% vs. 90% of face occluded), components (e.g. lens and frame), materials (e.g. glass or plastic), colors (e.g. grey or green), transparency (e.g. 20% or 50%), and texture, as well as temporal changes in these properties in the occluded region should be included in the annotation metadata is a question that is probably dependent on the aim of the database. These properties may potentially impact the performance of FEA methods. In the UMB 3D database [Colombo, et al. 2011], the type of object leading to the occlusion was annotated. In the HAPPEI database [Dhall *et al.* 2013], the intensity of facial occlusion for each person was manually annotated as one of three levels, including face visible, partial occlusion and high occlusion. The annotation can be used for evaluating the impact of different levels of occlusion on both individual- and group-level affect. Studies [Dhall *et al.* 2015a], [Dhall, et al. 2015b] have confirmed a big impact of occlusion on the perception and recognition of both levels of affect. Since the annotation provides just a rough category of the occlusion amount on the face, it may not be suitable for situations where an accurate percentage of occlusion is required. However, it is one of the earliest efforts towards the annotation of occlusion intensity in real-life images with multiple people. Studies [Cotter 2010a],[Zhang, et al. 2014b] have also shown that black and white occlusion lead to different classification accuracies, and occlusion of solid and clear eyeglasses also produce different results. Once the properties are decided, another issue arising is that how they can be properly annotated and saved in terms of ground truth labels and formats? One possible solution is to directly use raw pixels of occluded regions as ground truths, but this may not enable detailed analysis of specific features of the occlusion.

4) *Human labeling of emotion is a challenging and difficult task.* Human generally have no problem of recognizing a set of frequently occurring facial expressions (e.g., six basic emotions) from non-occluded clear faces in favorable conditions. However, they may face increased difficulty of recognizing the same type (or even more subtle and mixed types) of facial expression in the presence of obscured occlusion. Occlusion may significantly impact the accuracy and reliability of emotion labeling by humans. Once the most informative facial region was occluded, human labelers may find hard to correctly identify the predominant type of occlusion based on visual features in the remaining non-occluded parts. This problem may be relatively easy to be solved for labeling the occluded face into the most frequently prototypical categories of facial expressions, but may become challenging using uncommon emotion categories, FACS AUs, or continuous emotional dimensions. The AUs reflect the subtle and local muscle changes in local facial components. Once the most informative facial part for a certain AU is invisible due to occlusion, it is very difficult to accurately annotate the AU, and this may also impact the annotation of other AUs because many AUs are closely correlated and many emotions are represented via a combination of several AUs rather than a single AU. Emotion labeling using continuous dimensions requires well-trained labelers and detailed quantification of multiple dimensions based on subjective human perception. A simple solution is to leave the occlusion affected AUs or dimensions unlabeled. Another possible solution is to record both occluded and non-occluded facial expressions for each subject in the same recording settings, and use the emotion labeling in the non-occluded face as an estimation of the occluded face. Recent studies [Benitez-Quiroz, et al. 2016] have explored automatic annotation of AUs, AU intensities and emotion categories for a large number (> a million) of images of facial expressions collected from wild environments. However, the majority of the annotated images are unoccluded and facial occlusion was not specifically handled in these studies.

Aside from the above concerns associated with data creation and ground truth labeling, the size of the collected samples for each occlusion and each emotion, the closeness of the data to real-life scenarios, the time and expense costs, accessibility, construction and administration of the database are also important issues for consideration. The acquisition of 3D face data is also a critical step in motivating the investigation of 3D FEA models, particularly with the popularity of RGB-D cameras such as Microsoft Kinect. Whether other impacting factors such as pose variations and illumination changes should be jointly incorporated during the recording of facial occlusion is still a question that worthy considering. It is still arguable whether the occlusion should be imposed after or before facial expression recordings. Artificially imposed occlusion may solve some issues discussed above, but their capability of simulating real-life scenarios becomes a big concern.

## 6.2 Occlusion Detection and System Integration

Most current approaches extract features directly from occluded facial images without incorporating a pre-processing step of occlusion detection. They typically perform facial feature location, tracking, and extraction directly on the occluded face, or incorporate human assisted processing to manually crop the face and register facial landmarks.

For automatic FEA systems, the presence of occlusion may lead to imprecise facial feature localization, erroneous alignment or registration. The capacity of reliably determining the specific parameters of facial occlusion, such as the type, location, shape, appearance and temporal duration, forms a critical component of FEA systems. Once the parameters of occlusion were reliably measured or accurately determined, features in occluded parts can be either effectively reconstructed from training data based on prior knowledge of the face configuration or simply discarded from extracted features to minimize its effect on the performance. Prior knowledge about the parameters of occlusion has been proved as being crucial in boosting the performance of face recognition [Jongsun *et al.* 2005]. The benefit of incorporating occlusion detection as a pre-processing step in FEA systems have been demonstrated in terms of noticeable improved performance [Huang, et al. 2012].

Face occlusion detection and recovery is not a new field [Dahua and Xiaoou 2007], and tremendous efforts have been made towards the investigation of algorithms for robust face region detection [Burgos-Artizzu, *et al.* 2013],[Lin and Liu 2006], facial landmark localization [Ghiasi and Fowlkes 2014] and tracking [Torre *et al.* 2015], as well as face alignment [Heng *et al.* 2015], [Asthana *et al.* 2015] under partial occlusion. It is interesting to observe that recently developed face analyzers such as Intraface [Torre, *et al.* 2015] and incremental face model [Asthana *et al.* 2014] have achieved promising results of detecting, tracking and recognizing facial features even under moderate realistic occlusion. With those face analyzers, it is arguable that many types of temporal occlusion may not be a problem anymore, however, in our view, this progress nevertheless reduces the necessity and significance of designing occlusion detection techniques, which can provide details about the parameters and characteristics of occlusion to support more specific post-processing and analysis. The design of occlusion detectors is difficult primarily due to the random and varied characteristics of occlusion, and thus gaining a good understanding of the local context of the occlusion in a specific situation, such as the number of subjects, the type of occlusion, and the place (e.g. office or playground) becomes crucially important in simplifying the design process and to some extent, largely determines the performance of the algorithm. Developing an occlusion detector specific to a particular application might be a smarter choice than implementing a generic detector capable of detecting any possible type of occlusion in real scenarios.

## 6.3 Other Features

Existing approaches to handling facial occlusion for FEA are largely based on 2D gray data. There are relatively few efforts that specifically design 3D models to handle facial occlusion, and they are largely restricted to self-occlusion caused by pose variations. Very few studies have considered skin color features. For robust FEA with facial occlusion handling, it is desirable to adopt a richer set of representative and effective features.

1) *3D feature.* Facial data in 3D provides additional depth information about the facial structure and appearance on top of 2D image based facial features. It can be potentially utilized for generating more robust, discriminative, and view-independent features under facial variations, such as head pose, missing parts and partial occlusion [Drira, *et al.* 2013]. By giving insights into the comprehensive physical structure of the face, 3D features contain critical appearance and shape information for assisting occlusion detection and restoration [Colombo *et al.* 2010], facial landmark localization and recovery [Canavan *et al.* 2015],[Xi *et al.* 2011] and face alignment [Cao *et al.* 2014] in the presence of occlusion. They are also critically important in building a richer set of reliable features for representing facial expressions, and eventually leads to more accurate recognition. In the case that both occlusion and pose variations are present in the face region, 3D features can compensate the effect arising from pose movements by reconstructing the face to a frontal view [Kangkan *et al.* 2014], and simplify the problem to an occlusion only task. The generation of models using 3D facial features to handle other types of common occlusions such as a scarf, a mark, and glasses, is still a field that needs further investigation.

2) *Color feature.* Another valuable feature for reliable FEA under occlusion is color. Although occlusion may present in a form of varied colors, the skin color in the face can be roughly categorized into several big groups such as European, Asian and Hispanic. The skin color can be utilized as complementary information to assist the segmentation of occluded regions from the face and the extraction of a rich set of features. Skin color has already been successfully used to segment facial regions from complicated background objects [Ban *et al.* 2014] and detect occluded regions from the face [Lin, et al. 2013],[Lin and Liu 2006]. With respect to FEA, physiological studies [Nakajima *et al.* 2017] indicated that skin color is a useful clue for emotional states, for instance, the face often flushes during anger while goes pale for fear. Skin color also influences the perception of facial expressions by human. In addition, computer vision studies have also proved the usefulness of skin color for FEA. Ramirez *et al.* [2014] found that facial skin color is a reliable feature for inferring the valence of emotional states using machine learning algorithms. Studies [Lajevardi and Hong Ren 2012] have shown that skin color components convey additional features in achieving more robust and effective recognition of facial expressions in images with low-resolution or illumination variations. However, to our best knowledge, there is not work yet that directly utilized skin color features to handle facial occlusion for FEA. Thus, a possible future direction in this field is to explore the ways of using color features to detect and recover facial occlusion.

3) *Temporal feature.* Temporally dynamic features reflecting subtle or sudden spatio-temporal facial muscle movements in video sequences are also important for accurate FEA. Human facial expressions involve complex facial muscle interactions in both the space and time domains [Ziheng *et al.* 2013], and facial component and head motions can provide crucial information in assisting the human recognition of certain facial expressions [Bassili 1979],[Nusseck, et al. 2008]. For systematic occlusion (e.g. sunglasses) whose location is often roughly fixed in a certain region of the face, it is relatively easy to extract the dynamic movements of facial features in non-occluded facial parts. For temporary occlusion (e.g. hands moving across the face) whose location is usually varied, the missing facial features due to occlusion in a current frame can be recovered using temporal correlation reasoning on information in neighboring frames [Yongmian and Qiang 2005],[Miyakoshi and Kato 2011].

## 6.4 Multiple Modalities and Deep Learning

For occlusion-robust emotion analysis, the structure of the FER system can be extended from combining multiple modalities (going wider), exploiting multiple layers in a deep architecture (going deeper), or fusing both of them (going wider and deeper). Nearly all of existing studies on FEA under occlusion focus on visual features from the face only, and are limited to using a single deep learning architecture. It is anticipated that building a wider and/or deeper structure can lead to more robust performance.

1) *Multiple modalities.* It is still an unexplored field that exploits temporal correlations between multiple modalities and incorporates fused features from them in combating facial occlusion towards FEA. Human expression of emotions is often the result of interaction and collaboration between multiple modalities of human reactions, such as emotional voice, facial expressions, body gestures, head and shoulder movements, gaze direction, and physiological signs. Fused features from audio, visual, text, or physiological modalities have been extensively used for emotion analysis, yielding boosted performance compared to using a single modality alone [Zeng, *et al.* 2009],[Calvo and D'Mello 2010]. The great advantage of incorporating multiple modalities is that features from these modalities can be fully utilized to compensate the drawbacks of each other to generate more robust methods for handling occlusion. Because facial occlusion mainly impacts visual features in the face and normally has limited impact on audio, body gestures and physiological signals, integrating features in less-impacted modalities with facial features is anticipated to be able to identify emotions more robustly from an occluded face.

Several important issues need to be considered carefully for the fusion of multiple modalities [Zeng, *et al.* 2009], such as the selection of reliable modalities, the synchronization between signals with different characteristics (e.g. time scale, metric level, and temporal structure), extraction of discriminative temporal features from the raw signals, construction of joint features from multiple modalities, and fusion of classification decisions. Recent studies [Mahmoud *et al.* 2014] on the classification of hand-over-face gesture cues in naturalistic facial expression video are good examples towards robust emotion recognition using fusion of hand gestures and facial expressions under hand occlusion.

2) *Deep learning.* As reviewed in Section 4.6, current FEA studies on utilizing deep learning to handling facial occlusion are largely restricted to a single deep architecture. The potential capacity of adopting a wider and/or deeper structure using multiple data modalities has been evidenced as being the winners of the Wild Challenge and Workshop (EmotiW) from 2013 to 2016 [Kahou *et al.* 2013], [Liu *et al.* 2014a], [Fan *et al.* 2016]. Among them, Kahou *et al.* [2013] integrated four deep neural networks in a unique system for emotion recognition in video by capturing facial expressions, audio information, spatio-temporal patterns of human actions, and features in the mouth region. Liu *et al.* [Liu, et al. 2014a] mapped hand-crafted HOG, dense SIFT features, and CNN features into Riemannian manifolds and adopted a score-level fusion of three visual classifiers and an audio predictor for emotion recognition. Fan *et al.* [Fan, et al. 2016] presented a hybrid network that fuses RNNs, 3D convolutional networks, and an SVM based audio system in a late-fusion fashion for extracting appearance in individual frames, motion between frames, and audio information. It is noted that facial occlusion was not specifically handled in these work. However, it is anticipated that there will be a growing number of studies on investigating more complicated deep learning structures for handling occlusion in FEA.

## 6.5 A Few Additional Issues

1) *Multi-disciplinary experiment.* FEA is an inherently multi-disciplinary field and its progress is predominantly dependent on supports, knowledge and advance in closely related fields, including psychology, cognitive science, psychiatry and computer science. As the performance of contemporary machine systems is still far behind the innate recognition capability of the human, further investigations on the mechanisms of humans' recognition behaviors in the presence of facial occlusion may be a crucially important step in gaining invaluable insights and relevant knowledge that can potentially inspire the way of designing reliable FEA systems. For instance, it is still not fully understood whether humans primarily adopt holistic or component-based strategies for recognizing expressions from partially occluded faces, and how the human brain instantaneously recovers features obscured by occlusion. Studies [Hammal, et al. 2009] revealed that the human tends to use "suboptimal" features for FEA under occlusion. It is advisable to conduct both computer vision and human perception experiments on the same type of occlusion and the same dataset so that the results can be directly comparable and novel insights can be obtained. These insights can be used, for instance, to focus on extracting features from the most important facial parts for a specific expression.

2) *Context.* As reviewed in Section 5.3, the effect of partial occlusion on the classification of facial expressions is largely context-dependent. The local environment (i.e. context) in which facial occlusions are imposed, facial expressions are elicited and simultaneously recorded may have big impact on the collected expressions in terms of spontaneity and exaggeration levels. The context also carries prior knowledge about specific parameters (e.g. type, location, appearance and time) of the occlusion that is going to occur, which is critically important for all procedures of occlusion detection, feature extraction, and emotion classification in a FEA system specifically designed for handling this type of occlusion. The contextual information may include the place, time, surrounding people and background in the recording environment, as well as the subject's personal backgrounds, such as the age, gender, job, culture and habit. Recent studies [Rudovic *et al.* 2015] on context-sensitive modelling of the AU intensity with respect to six context questions (who, when, what, where, why and how) achieve substantially improved accuracies than without considering the effects of the context. The incremental face model [Asthana, et al. 2014], which is capable of automatically tailor itself to fit specific person and imaging conditions, has shown accurate face tracking even under temporal occlusion, fast head movement, shadow, and pose variation.

3) Group-level expression analysis. Recently, increasing attention has been given to group-level expression analysis, which aims to identify the type and intensity of emotions from images of a group of people. Aside from ordinary occlusion such as sunglasses, hat and beard, one frequently occurring occlusion in the images is partial facial occlusion due to the presence of another person standing in front of the face. The presence of these occlusion was found as one of the key attributes that affects the perception of the emotion of a group [Dhall, et al. 2015b]. It has also been shown that people tend to select images with less occlusion of the face in the process of identifying happiness intensity of a group [Dhall *et al.* 2015a]. One advantage of group-level expression analysis is that, even the face of one (or more) person is partially occluded, the dominant emotion of the group still can be inferred by fusing facial information of all group members in conjunction with the holistic scene context. Such a 'fusion' strategy provides another angle of handling facial occlusion in real-life scenarios, and has been adopted in recent methods for group-level expression analysis, particularly in the EmotiW 2016. The fusion can be generally either in feature-level or decision-level. In [Li *et al.* 2016], holistic features from the whole image scene and local features from multiple faces were learnt using a ResNet-18, and further aggregated to a feature vector using a LSTM. The aggregated feature vector was fed to linear or ordinal regression to predict group-level happiness intensities. In [Huang *et al.* 2015], Riesz-based Volume LBP features were extracted to represent the local attribute of each face, and relative sizes and distances between all faces were used to represent the global attribute of the scene. They were concatenated and fed to continuous conditional random fields for predicting the group mood. As for decision-level fusion, [Dhall, et al. 2015b] extracted Bag of Words representations of different features from multiple faces and GIST and census transform histogram features from the scene. Each feature modality was considered as a separate kernel and all kernels were linearly combined using a Multiple Kernel Learning (MKL) to predict happiness intensity. In [Sun *et al.* 2016], facial features were learnt by applying CNN and LSTM consequently to the detected face region. The features of each face were used to train an individual SVM classifier, and the prediction results of all SVMs were combined using a decision-level fusion network. Although the above works were not designed specifically to handle facial occlusion, the occluded part was inherently considered by the feature extraction or fusion techniques such as CNN, LSTM, and MKL.

4) *International competition.* The international competitions for FEA under non-occluded faces have been very active in recent years, such as the FERA 2011, 2015 and 2017 challenges [Valstar *et al.* 2015], [Valstar, et al. 2017], which focus on expression recognition using discrete emotion categories and estimating the occurrence and intensity of AUs, the Audio/Visual Emotion Challenge and Workshop (AVEC) 2011-2016 [Ringeval *et al.* 2015],[Valstar *et al.* 2016], which focus on using continuous dimension representations, and the EmotiW 2013-2016 [Dhall *et al.* 2015c], [Dhall *et al.* 2016a] which focus on using data collected from the wild. With recent studies on FEA have shifted towards using continuous representations [Kaltwang *et al.* 2015] and spontaneous emotions in a wild environment for practical applications [Zhang, et al. 2014a],[Zhang *et al.* 2016], there are still very few international competitions that are specifically designed to compare FEA systems with partially occluded face data. The EmotiW 2016 includes a group-level emotion recognition sub-challenge, which aims to compare methods for predicting the happiness intensity of a group of people. The benchmark HAPPEI database covers various types of realistic occlusion (e.g., sunglasses, hat, and a people standing in front of another and partially occluding the face) and meta-data of three intensities of facial occlusion, which makes it possible to evaluate the effect of occlusion on the perception and recognition of emotions of a group [Dhall, et al. 2015a]. The most recent EmotiW 2017 has incorporated images with similar occlusion and three emotions (e.g., positive, negative and neutral) from the Group Affect database. Another recent international competition - FERA 2017 has also started to consider self-occlusion caused by pose variations. It is anticipated that joint initiatives similar to FERA, AVEC and EmotiW will greatly promote the research in this direction by providing a common platform for performance evaluations, such as benchmark datasets, predefined targets, train-test guidelines, evaluation procedures, performance measures, and baseline approaches.

## 7 Conclusion

This paper presents a survey on the state-of-the-art efforts and a discussion about relevant challenges and opportunities for handling partial occlusion towards automatic Facial Expression Analysis (FEA). In the last decade, while an increased amount of studies have been recorded on handling occlusion, most FEA systems capable of overcoming occlusion are still at the early stage, characterized by a very limited number of prototypical emotion categories and artificially generated occlusion. Features are completely restricted to the visual face modality only, and evaluations are primarily based on 2D/3D frontal faces. Amongst all types of existing approaches reviewed, the sparse representation and deep learning approaches have demonstrated the most impressive results in combating facial occlusion.

Existing studies on FEA under partial occlusion still lack of:

- comprehensive benchmark datasets that include a dense set of various types of frequent natural facial occlusion and well annotated ground truths of facial expressions by not only discrete categories, but also AUs and dimensional axes;

- work on designing face occlusion detection techniques to reliably determine the specific parameters of facial occlusion, such as the type and location;
- investigations on exploiting temporal correlations between multiple modalities and incorporating fused features from multiple modalities in combating facial occlusion;
- efforts on thoroughly investigating the effect of facial occlusion on the performance of non-prototypical spontaneous emotions across multiple realistic datasets.

Future FEA systems in handling facial occlusion are expected to expand from:
- artificially imposed to naturally occurring occlusion;
- 2D to 3D face databases;
- manual face pre-processing to automatic occlusion detection and integration;
- static 2D grey to temporal 3D color features;
- a single face to multiple faces of a group of people;
- a single face modality to multiple audio, visual and physiological modalities;
- a shallow architecture to deeper and wider architectures;
- prototypical emotions to AU-coded, continuously represented emotions, and micro-expressions.

With the emergence of more comprehensive benchmark datasets and the launch of international joint efforts and initiatives, new algorithms will be developed subsequently, eventually leading to automated machine systems that can support FEA applications in unconstrained conditions including the presence of occlusion. As a multi-discipline field, FEA can benefit substantially from advancement in the knowledge in closely related areas of computer science, psychology, cognitive science, neuroscience, etc. Occlusion as a challenge is not specific to FEA, but also exists in relevant fields such as face recognition, face detection and face tracking. Studies in these fields share many common techniques, knowledge, issues, and challenges, and thus any progress in one field can potentially benefit to other fields. A promising direction for further research is the development of context-sensitive FEA algorithms that take into account the prior knowledge of the local environment to predict the specific parameters of facial occlusion. It is also worthy conducting further investigations on the power of deep learning techniques in recovering and handling facial occlusion inherently without human manual intervention.